\documentclass{bmvc2k}

\title{Context-Guided Semantic Alignment\\for Feature Fusion Networks}

\addauthor{Hyungseop Lee}{hhss0927@inu.ac.kr}{1}
\addauthor{Jiho Lee}{jiho264@inu.ac.kr}{1}
\addauthor{Woochul Kang$^*$}{wchkang@inu.ac.kr}{1}

\addinstitution{
 Department of Embedded Systems Engineering\\
 Incheon National University\\
 Yeonsu-gu, Incheon, South Korea, 22012
}

\runninghead{H. Lee, J. Lee, W. Kang}{Context-Guided Semantic Alignment}

\def\eg{\emph{e.g}\bmvaOneDot}

\usepackage{booktabs}
\usepackage{multirow}
\usepackage{colortbl}
\usepackage{array}
\usepackage{wrapfig}
\usepackage{amsmath}
\usepackage{enumitem}
\usepackage{hyperref}
\usepackage{amsfonts}
\usepackage{algorithm}
\usepackage{listings}
\usepackage{xcolor}

\input{glyphtounicode}
\begin{document}

\maketitle

\begingroup
\renewcommand\thefootnote{}
\footnotetext{\hspace{-1.8em}$^*$ Corresponding author.}
\endgroup


\begin{abstract}
Feature fusion networks are fundamental components in modern object detectors, aggregating multi-scale features to detect objects of varying sizes. 
However, directly fusing features from different pyramid levels often introduces semantic inconsistency, causing information conflicts that distort the fused representation and degrade detection accuracy.
In this paper, we propose Feature Interaction NEtwork (FINE), a lightweight semantic alignment module that refines low-level features via high-level contextual guidance using cross-level attention prior to fusion. 
To bridge the structural gap and ensure computational efficiency, we introduce an Alignment-Aware Token Sampling that aligns corresponding spatial regions across scales, reducing the attention complexity by an order of magnitude.
The resulting attention weights generate a spatial-channel modulation map that is applied to the low-level features via residual modulation. 
This mechanism ensures that the network selectively enhances semantically relevant pixels while preserving the sub-pixel localization accuracy necessary for dense prediction tasks.
FINE is generally applicable to various detectors and consistently improves detection accuracy with minimal computational overhead.
Our code is publicly available at \url{https://github.com/HyungseopLee/FINE}.
\end{abstract}
\section{Introduction}
\label{sec:intro}

\begin{figure*}[!t]
\centering
\includegraphics[width=0.95\linewidth]{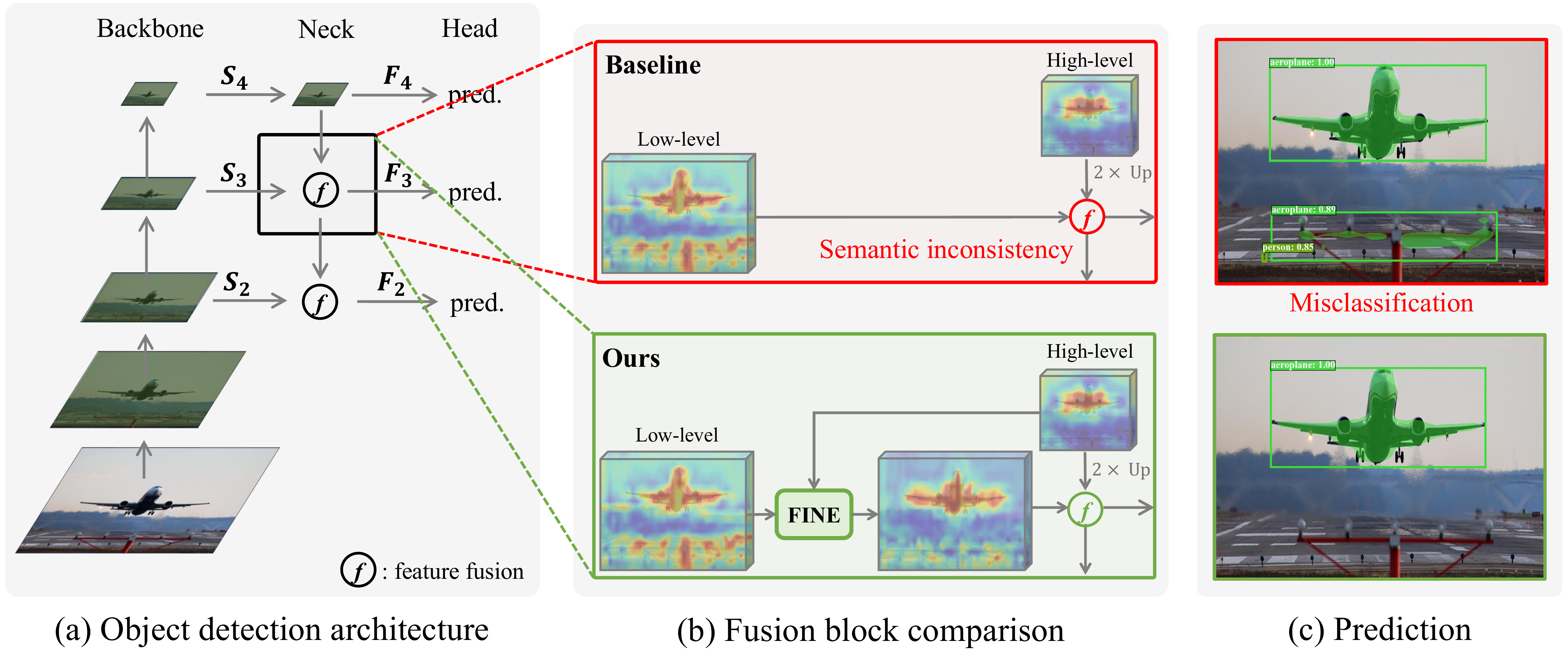}
\caption{
    \textbf{Semantic inconsistency in conventional feature fusion.}
    \textbf{(a)} Modern detectors use a neck to aggregate multi-scale backbone features.
    \textbf{(b)} Standard fusion blocks combine features via naive element-wise operations, enforcing pixel-wise alignment but leaving representational discrepancies unresolved. 
    In contrast, our method mitigates this by explicitly aligning low-level features with high-level context prior to fusion.
    \textbf{(c)} In baseline models, semantic inconsistency often causes severe detection errors, such as background misclassifications or missed objects. 
    FINE reduces such errors and improves detection reliability.
}
\label{fig:fig1}
\end{figure*}

Modern object detectors~\cite{yolov4, yolov10, yolov12, RTDETRv1, RTDETRv3, DEIM} generally follow a tripartite architecture, consisting of a backbone for feature extraction, a neck for feature fusion, and a head for prediction, as illustrated in Figure~\ref{fig:fig1}(a).
The backbone extracts hierarchical representations from the input image, where early layers capture fine-grained local details with small receptive fields, while deeper layers produce semantically rich but spatially coarse representations due to progressively expanded receptive fields.
To exploit complementary information across feature levels, the neck aggregates these pyramidal features via cross-level fusion, and the fused representations are fed into the head for bounding box regression and category classification.
As an intermediate component bridging feature extraction and final prediction, the neck is responsible for constructing multi-scale representations essential for detecting objects of varying sizes.

Despite the widespread adoption of feature fusion networks~\cite{FPN, PANet, NAS-FPN, BiFPN}, existing approaches often overlook a critical issue: \textit{semantic inconsistency} between feature levels.
In typical feature pyramid architectures, high-level features learn rich semantic contexts while losing spatial information due to repeated downsampling in the backbone network.
In contrast, low-level features preserve spatial details but lack sufficient semantic abstraction.
Directly fusing these heterogeneous features through naive element-wise operations, such as addition or concatenation, enforces position-wise aggregation without resolving the underlying representational discrepancies, as illustrated by the activation heatmaps in Figure~\ref{fig:fig1}(b).
This semantic mismatch induces information conflicts~\cite{A2-FPN} at corresponding spatial positions, thereby distorting the fused representation and leading to severe detection errors, as shown in Figure~\ref{fig:fig1}(c) (detailed analysis in Section~\ref{subsec:exp_analysis}).

Motivated by these limitations, we propose a lightweight semantic alignment module, termed \textbf{F}eature \textbf{I}nteraction \textbf{NE}twork (FINE).
Before feature fusion, FINE performs spatial-channel modulation on low-level features, guided by the rich context of their corresponding higher-level representations.
To generate this modulation map, we employ multi-head cross-level attention to capture diverse semantic interactions. 
However, directly enforcing dense pixel-to-pixel interactions across pyramid levels is not only computationally prohibitive but also inherently unaligned in terms of effective receptive fields.
Specifically, features from different backbone stages encode distinct contextual scales and exhibit inconsistent receptive field sizes, making naive position-wise interactions unreliable for establishing meaningful cross-level correspondences.
To address this, we introduce Alignment-Aware Token Sampling (AATS), which condenses features into representative tokens while aligning their effective receptive fields across levels. 
Based on these aligned and reduced tokens, our bottleneck cross-level attention establishes robust correspondences to selectively inject high-level semantic context into low-level features.
Since the modulation map is estimated at a coarse regional scale due to token sampling, we adopt a residual modulation strategy to preserve high-resolution spatial cues critical for accurate object localization.

Extensive experiments on the MS COCO~\cite{COCO} dataset demonstrate that FINE consistently boosts performance across various detection architectures with negligible computational overhead.
Furthermore, FINE yields notable gains for small object detection by enriching the limited semantic context of low-level features.
While our evaluation primarily focuses on object detection on MS COCO, FINE can be integrated into other dense prediction tasks and generalizes to other datasets, as detailed in Appendix Sections~\ref{appen_subsec:other_dense} and~\ref{appendix_sec:VisDrone}.

Our contributions can be summarized as follows:
\begin{itemize}
\item 
We propose FINE, a lightweight plug-and-play module that enables efficient cross-level attention to mitigate semantic inconsistency across feature pyramid levels in real-time detectors.

\item 
We propose Alignment-Aware Token Sampling, a mechanism that significantly reduces the computational overhead of cross-level attention while establishing precise regional correspondences across pyramidal feature levels.


\item 
Extensive evaluations demonstrate that FINE consistently enhances detection performance across diverse architectures with minimal inference overhead, yielding particularly significant improvements in small object detection.
\end{itemize}

\section{Related Work}
\label{sec:related}

\subsection{Misalignment in Feature Fusion Networks}

Semantic misalignment across pyramid levels poses a fundamental challenge in feature fusion networks. 
To alleviate this issue, several approaches recalibrate channel-wise feature responses.
For example, Squeeze-and-Excitation (SENet)~\cite{SENet} is widely used to modulate channel importance. 
Building on this, FaPN~\cite{FaPN} emphasizes informative channels via a feature selection module, while AFF~\cite{AFF} extends SENet with multi-scale pooling for global and local context.
AugFPN~\cite{AugFPN} further mitigates semantic gaps by enforcing consistent supervision across all pyramid levels. 
Other works redesign the fusion architecture: for instance, $A^2$-FPN~\cite{A2-FPN} aggregates global context from multiple levels to reduce semantic loss, and CATFPN~\cite{CATFPN} refines separate feature pyramids using global context blocks. 
In parallel, some studies address pixel-level misalignment caused by artifacts from naive upsampling, employing learnable sampling offsets~\cite{FaPN, AdaFPN, AlignSeg, DeformableAttentionOrientedFPN} or soft upsampling~\cite{SNI, EffectiveFF} to preserve fine-grained textures.
While these approaches improve fusion quality, they often introduce significant computational overhead or lack plug-and-play applicability to existing detectors.

\subsection{Attention Mechanisms for Object Recognition}

Self-attention~\cite{AttentionIsAllYouNeed} has been widely adopted to model long-range dependencies.
However, its quadratic complexity \cite{ViT} has motivated numerous efficient variants, 
including local attention \cite{SwinT, Criss-Cross, yolov10, yolov12}, 
hybrid CNN--transformer architectures \cite{CVT, MobileViT}, 
kernel-based linear attention \cite{SOFT, EfficientViT, SimA}, 
and token sparsification \cite{DynamicViT, ToMe, EViT}.
Despite their efficacy, these mechanisms are tailored for intra-level feature extraction within a single scale.

Several studies have leveraged cross-attention to model dependencies across heterogeneous feature representations.
For multi-scale feature fusion, FPT \cite{FPT} introduces transformer-style interactions across the feature pyramid, 
while CLASS \cite{CLASS} employs cross-level attention in a U-Net-based \cite{UNet} architecture for salient object detection.
However, their dense pixel-wise computations introduce substantial latency, limiting their applicability to real-time architectures.
While the DETR family \cite{DETR, DeformableDETR, DN-DETR, DAB-DETR} and its real-time variants \cite{RTDETRv1, RTDETRv3, DEIM, D-FINE} employ cross-attention for generic object detection, this mechanism is primarily used in the detection head for decoding object queries, rather than for cross-level feature fusion in the neck. 
Notably, SAM-DETR~\cite{SAM-DETR} is closely related to our work in addressing semantic misalignment, but it focuses on resolving representational mismatches between encoder features and queries for cross-attention in the decoder, rather than cross-level feature fusion.
Consequently, efficient cross-level attention within feature fusion networks remains largely unexplored in real-time object detection, hindered by the prohibitive computational costs of high-resolution features and the spatial-channel heterogeneity.






\section{Preliminaries}
\label{sec:pre}

In this section, we briefly review Feature Pyramid Networks (FPN)~\cite{FPN} and establish the notation used throughout this paper.
Given an input image, a backbone network extracts a hierarchy of feature maps $\{S_2, S_3, S_4\}$, where each $S_l \in \mathbb{R}^{H_l \times W_l \times C_l}$ corresponds to strides of $\{8, 16, 32\}$ pixels relative to the input image, respectively. 
Through lateral connections implemented as $1 \times 1$ convolutions, each $S_l$ is first projected to a unified channel dimension $C$. 
At the $l$-th fusion stage, the low-level and high-level features are defined as:

\begin{equation}
    F_{\text{low}} = \text{Conv}_{1\times1}(S_l),
    \quad
    F_{\text{high}} = \text{Conv}_{1\times1}(S_{l+1}).
\label{eq:1x1Conv}
\end{equation}
These adjacent features are integrated using a top-down pathway:
\begin{equation}
    F_{\text{fused}} = f\big(F_{\text{low}}, \text{Up}(F_{\text{high}}, \text{scale}=2)\big),
\label{eq:naive_fusion}
\end{equation}
where $\text{Up}(\cdot)$ denotes nearest-neighbor interpolation and $f(\cdot)$ represents a fusion operator, such as element-wise addition or concatenation.
The resulting $F_{\text{fused}}$ subsequently serves as the high-level input for the next fusion stage ($l-1$). 
Through this recursive process, the network constructs multi-scale features $\{F_2, F_3, F_4\}$, integrating fine-grained spatial details with high-level semantic information. 
These features may be further refined by an additional pathway~\cite{PANet} or fed directly into the detection head.

\section{Method}
\label{sec:method}
\subsection{Overview}
\label{sec:overview}

\begin{figure*}[!t]
\centering
\includegraphics[width=0.9\textwidth]{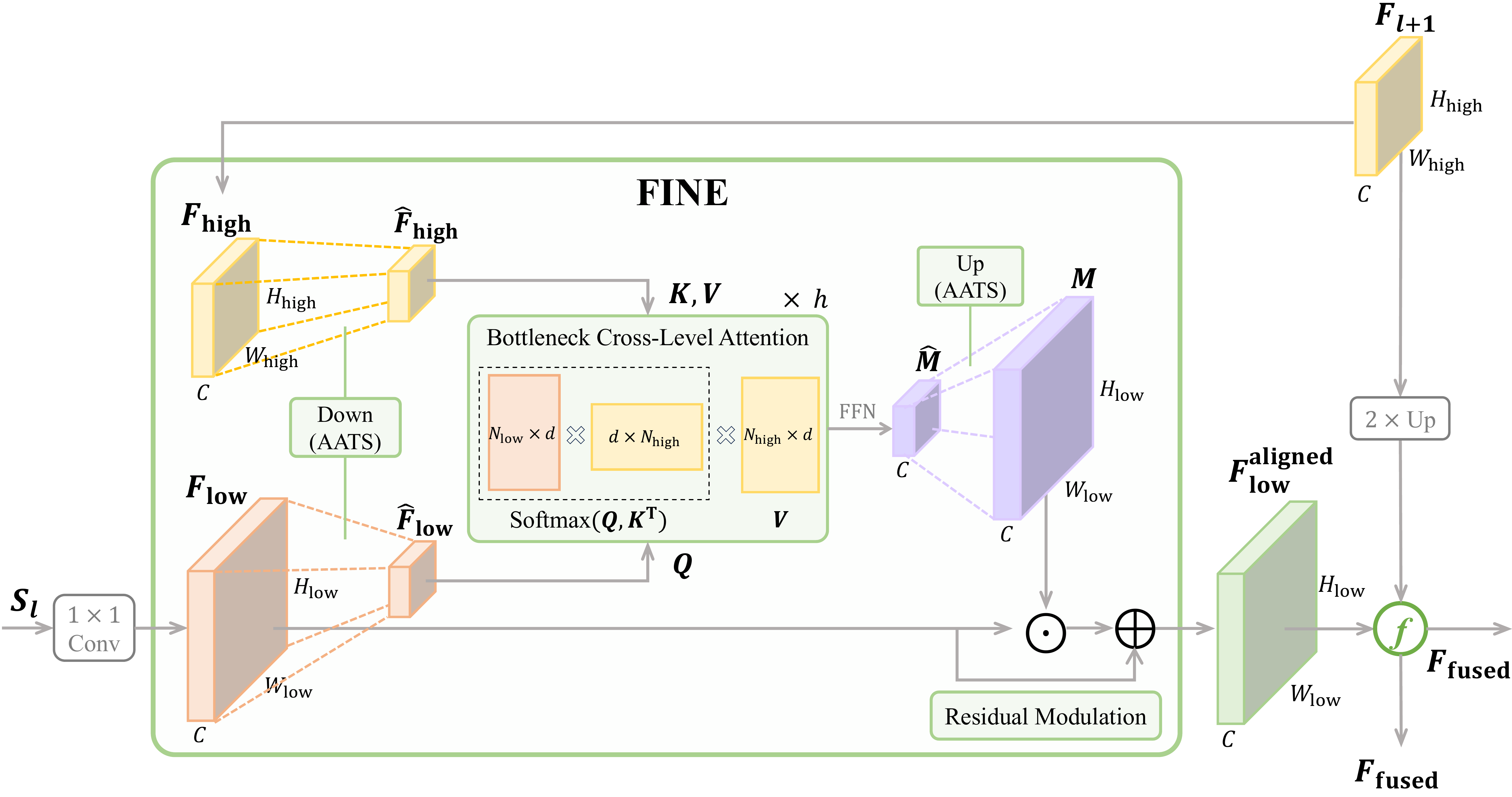} 
\caption{
    \textbf{Overview of the Feature Interaction NEtwork (FINE).}
    Given adjacent low- and high-level features ($F_{\text{low}}$ and $F_{\text{high}}$), 
    FINE refines $F_{\text{low}}$ into a semantically aligned representation $F_{\text{low}}^{\text{aligned}}$ under the guidance of $F_{\text{high}}$. 
    Through Alignment-Aware Token Sampling (AATS), $F_{\text{low}}$ is condensed into $\hat{F}_{\text{low}}$ as the Query ($Q$), and $F_{\text{high}}$ is compressed into $\hat{F}_{\text{high}}$ as the Key ($K$) and Value ($V$), 
    with an alignment factor that explicitly aligns their regional correspondence.
    The resulting bottleneck cross-level attention produces a spatial-channel modulation map $\hat{M}$.
    After upsampling, the map is applied to $F_{\text{low}}$ via element-wise multiplication, followed by a residual connection to preserve fine-grained localization cues.
}
\label{fig:overview_FINE}
\end{figure*}

Figure~\ref{fig:overview_FINE} illustrates an overview of our \textbf{F}eature \textbf{I}nteraction \textbf{NE}twork (FINE).
We reformulate the standard fusion process Eq.~\ref{eq:naive_fusion} by replacing the original low-level feature $F_{\text{low}}$ with its semantically aligned representation $F_{\text{low}}^{\text{aligned}}$, conditioned on the high-level context $F_{\text{high}}$:
\begin{equation}
    \begin{aligned}
        F_{\text{fused}} &= f\big(F_{\text{low}}^{\text{aligned}}, \text{Up}(F_{\text{high}}, \text{scale=2})\big), \\
        \text{where } F_{\text{low}}^{\text{aligned}} &= \text{FINE}(F_{\text{low}}, F_{\text{high}}).
    \end{aligned}
\label{eq:our_fusion}
\end{equation}

To generate $F_{\text{low}}^{\text{aligned}}$, the FINE module employs three key components:
(i) Alignment-Aware Token Sampling (Section~\ref{subsec:AATS}) to align inherently unaligned effective receptive fields and reduce memory and computational cost, 
(ii) Bottleneck Multi-Head Cross-Level Attention (Section~\ref{subsec:BMCA}) to construct a spatial-channel modulation map that captures diverse cross-level semantic interactions across multiple subspaces, and 
(iii) Residual Spatial-Channel Modulation (Section~\ref{subsec:residual_modulation}) to apply the estimated modulation, selectively injecting high-level context into low-level features while preserving fine spatial details.


\subsection{Alignment-Aware Token Sampling}
\label{subsec:AATS}

\begin{figure*}[!t]
\centering
\includegraphics[width=0.95\textwidth]{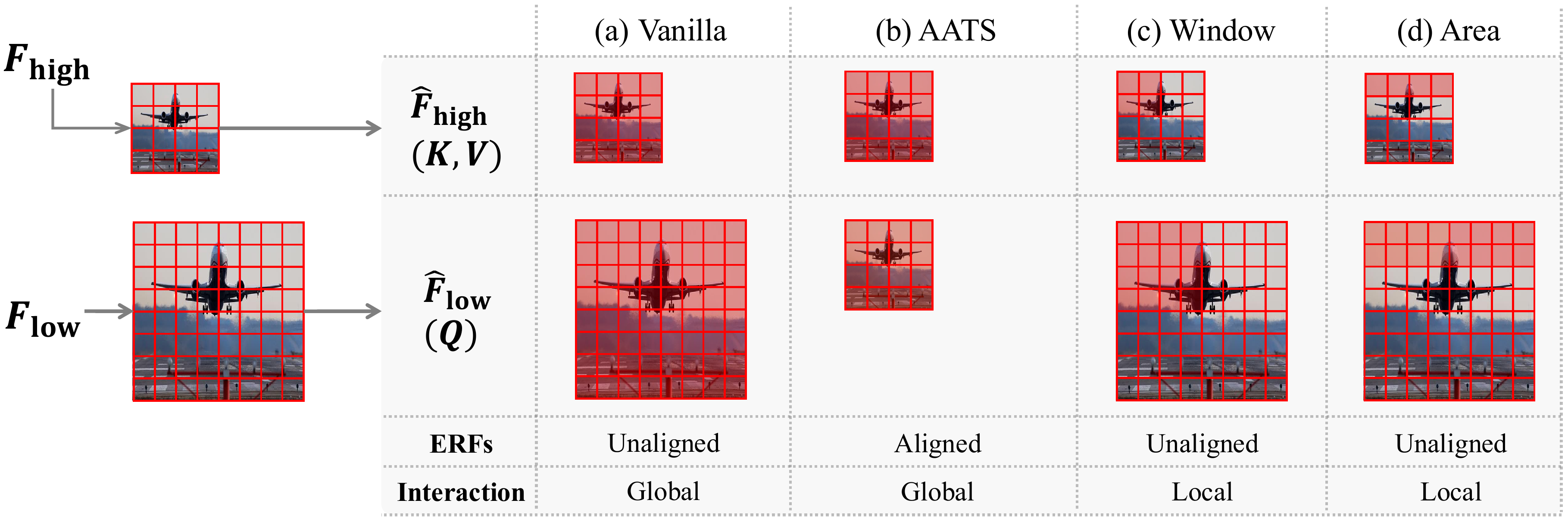} 
\caption{
    \textbf{Visualization of cross-level token interactions.}
    Each grid cell represents a discrete token within its respective feature map, illustrating the scale disparity between adjacent levels.
    Shaded grids indicate tokens participating in attention.
    \textbf{(a)} In the vanilla mechanism, naive pixel-to-pixel matching pairs tokens that cover different spatial scales due to the varying strides of the backbone.
    \textbf{(b)} Alignment-Aware Token Sampling (AATS) structurally aligns the representation scales across levels, enabling reliable cross-level correspondences for accurate global interaction.
    Local attention variants including \textbf{(c)} window and \textbf{(d)} area attention restrict long-range dependencies and leave the scale mismatch unresolved.
}
\label{fig:AATS}
\end{figure*}

To alleviate semantic inconsistency across adjacent pyramid levels, we employ 
cross-level attention to refine the low-level feature $F_{\text{low}}$ under the 
semantic guidance of the high-level feature $F_{\text{high}}$. 
However, vanilla dense pixel-to-pixel interaction, as visualized in Figure~\ref{fig:AATS}(a), faces a fundamental limitation in pyramidal architectures: features from adjacent backbone stages inherently possess different effective receptive fields (ERFs)~\cite{ERF}, \textit{i.e.}, the spatial region of the input image that influences each token.
Since $F_{\text{high}}$ is extracted at a spatial stride $s$ times larger than that of $F_{\text{low}}$ ($s > 1$, typically $s=2$), tokens from different stages correspond to inconsistent physical regions of the input image.
To resolve this, we structurally synchronize the spatial coverage across levels, establishing reliable cross-level correspondence for accurate global interaction, as depicted in Figure~\ref{fig:AATS}(b).
Visualizations of the actual ERF distributions are provided in Figure~\ref{fig:ERF_heatmap_paired}.

Furthermore, the massive token count in high-resolution visual features makes the vanilla approach computationally prohibitive for real-time applications.
While lightweight variants such as window-~\cite{SwinT} or area-based schemes~\cite{yolov12}, as depicted in Figure~\ref{fig:AATS}(c,d), improve efficiency by restricting attention to local regions, they limit the long-range dependencies of $F_{\text{high}}$ and fail to align the ERFs.

To establish ERF alignment and mitigate the excessive token density, we introduce an \textit{Alignment-Aware Token Sampling} (AATS) strategy that constructs compact, ERF-aligned token representations prior to cross-level attention. 
AATS downsamples both feature levels while accounting for their stride discrepancy, using asymmetric sampling kernels to align their effective receptive fields:
\begin{equation}
\begin{aligned}
    \hat{F}_{\text{high}} &= \text{Down}(F_{\text{high}}, \text{kernel}=k), \\
    \hat{F}_{\text{low}} &= \text{Down}(F_{\text{low}}, \text{kernel}=rk)
\end{aligned}
\label{eq:AATS}
\end{equation}
where $k$ is the base sampling kernel size for the high-level feature, and $r$ is the \textit{alignment-aware sampling ratio}.

Assuming non-overlapping pooling, the resulting features are 
$\hat{F}_{\text{high}} \in \mathbb{R}^{\frac{H_{\text{high}}}{k} \times \frac{W_{\text{high}}}{k} \times C}$ and 
$\hat{F}_{\text{low}} \in \mathbb{R}^{\frac{H_{\text{low}}}{rk} \times \frac{W_{\text{low}}}{rk} \times C}$. 
Since the choice of pooling operator has marginal impact on cross-level attention quality, we adopt parameter-free average pooling as our default implementation of AATS (see Appendix Section~\ref{appen_subsec:downsampling} for an ablation study).

For all object detectors in the evaluation, we set the sampling ratio $r=s=2$ to match the downsampling ratio between adjacent backbone stages.
Empirical validation of this choice is provided in Section~\ref{subsubsec:align_aware_samplnig_ratio}, and a theoretical rationale in terms of ERF alignment is given in Appendix Section~\ref{appendix_sec:ERF}.

\subsection{Bottleneck Multi-Head Cross-Level Attention}
\label{subsec:BMCA}

Given the ERF-aligned, condensed feature maps $\hat{F}_{\text{low}}$ and
$\hat{F}_{\text{high}}$ from AATS, 
we apply multi-head cross-level attention in a bottleneck manner to generate a spatial-channel modulation map $M$.
By assigning $\hat{F}_{\text{low}}$ as the query and $\hat{F}_{\text{high}}$ as the key and value, each spatial location in $\hat{F}_{\text{low}}$ selectively retrieves semantically relevant contextual information from higher-level representations.
The condensed maps are flattened and linearly projected to obtain queries ($Q$), keys ($K$), and values ($V$):
\begin{equation}
    Q = \text{Flatten}(\hat{F}_{\text{low}})W_Q, \quad 
    K = \text{Flatten}(\hat{F}_{\text{high}})W_K, \quad 
    V = \text{Flatten}(\hat{F}_{\text{high}})W_V,
\label{eq:QKV_embeddings}
\end{equation}
where $W_Q, W_K, W_V \in \mathbb{R}^{C \times C}$ are learnable projection weights. 
This yields $Q \in \mathbb{R}^{N_{\text{low}} \times C}$ and $K, V \in \mathbb{R}^{N_{\text{high}} \times C}$, with reduced sequence lengths $N_{\text{low}} = \frac{H_{\text{low}} W_{\text{low}}}{r^2 k^2}$ and $N_{\text{high}} = \frac{H_{\text{high}} W_{\text{high}}}{k^2}$.

To capture diverse semantic-spatial interactions across heterogeneous feature levels, we adopt multi-head attention.
By projecting $Q$, $K$, and $V$ into $h$ independent subspaces (each with dimension $d=C/h$), the network jointly captures cross-level correlations from multiple representation perspectives rather than a single shared space.
Formally, for the $i$-th head, the cross-level attention is computed as:
\begin{equation}
\text{head}_i = \text{Softmax}\!\left( \frac{Q_i K_i^\top}{\sqrt{d}} \right) V_i
\in \mathbb{R}^{N_{\text{low}} \times d},
\label{eq:each_head}
\end{equation}
where the attention weights identify cross-level token correspondences,
and $V_i$ carries the high-level semantic content to be selectively
injected into each low-level spatial location.
The outputs from all heads are concatenated and linearly projected. 
This attention output is further processed by a Feed-Forward Network (FFN) and reshaped into a spatial feature map $\hat{M}$:
\begin{equation}
    \hat{M} = \text{Reshape}\Big( \text{FFN}\big(\text{Concat}(\text{head}_1, \dots, \text{head}_h) W^O\big) \Big) \in \mathbb{R}^{\frac{H_{\text{low}}}{rk} \times \frac{W_{\text{low}}}{rk} \times C}.
\end{equation}
Finally, $\hat{M}$ is spatially upsampled by a factor of $rk$ to recover the original resolution of $F_{\text{low}}$:
\begin{equation}
    M = \text{Up}(\hat{M}, \text{scale=}rk) \in \mathbb{R}^{H_{\text{low}} \times W_{\text{low}} \times C}.
\end{equation}
This upsampling broadcasts the encoded semantic context across $rk \times rk$ local regions, providing spatially aligned guidance for subsequent modulation (Section~\ref{subsec:residual_modulation}).

\subsection{Residual Spatial-Channel Modulation}
\label{subsec:residual_modulation}

Instead of directly using the upsampled attention output $M$ as a standalone feature, we use it to refine the original low-level feature $F_{\text{low}}$. 
Since $M$ is estimated from condensed tokens via AATS, it is spatially coarser than the dense feature map, and directly replacing $F_{\text{low}}$ with $M$ would discard fine-grained details essential for precise dense prediction. 
We therefore treat $M$ as a spatial-channel modulation map that adaptively re-weights $F_{\text{low}}$ rather than overwriting it. 
Unlike conventional channel-attention methods that apply a single global scaling factor per channel, $M$ provides location-specific modulation across both spatial and channel dimensions, enabling the network to emphasize object-relevant regions while suppressing background noise.

However, direct modulation ($F_{\text{low}}^{\text{aligned}} = F_{\text{low}} \odot M$) tightly couples the refined features to the coarse spatial structure inherited from the bottleneck attention, which may attenuate fine-grained spatial details crucial for small object detection.
To provide a more conservative refinement pathway, we formulate the modulation as a residual transformation:
\begin{equation}
F_{\text{low}}^{\text{aligned}} = (F_{\text{low}} \odot M) + F_{\text{low}},
\label{eq:residual_modulation}
\end{equation}
where $\odot$ denotes element-wise multiplication. 
The identity mapping ($+F_{\text{low}}$) provides a direct information pathway, preserving the original high-resolution spatial details.
The modulation term $(F_{\text{low}} \odot M)$ functions as a data-dependent residual transformation that adaptively rescales feature responses based on high-level contextual cues.


\section{Experiments}
\label{sec:exp}

\subsection{Experimental Setup}

We conduct experiments on the MS COCO dataset~\cite{COCO}, which consists of 118k training images and 5k validation images.  
Following standard practice, we evaluate object detection performance using the official COCO metrics: mean Average Precision (AP).
For a fair comparison, FINE is plugged into each baseline without any modification to the original training recipes.
To evaluate efficiency under resource constraints, FPS and latency are measured on an NVIDIA Jetson Orin Nano with TensorRT v10.7.0~\cite{tensorrt}.

\begin{table*}[!t]
    \begin{center}
        \resizebox{0.95\textwidth}{!}{
            \begin{tabular}{ll | ccc | lccccc}
                \toprule
                Model & Fusion Method & \#Params & FLOPs & FPS & AP$_{50:95}$ & AP$_{50}$ & AP$_s$ & AP$_m$ & AP$_l$ & \\
                \midrule
                
                \rowcolor{gray!10}
                \multicolumn{11}{c}{SOTA Real-Time Object Detectors} \\
                \midrule
                RT-DETRv1 R18~\cite{RTDETRv1}      & H-Enc    & 20.2M   & 61.7G   & 80   & 46.5  & 63.8 & 28.4 & 49.8 & 63.0 & \\
                RT-DETRv1 R50$^*$~\cite{RTDETRv1} & H-Enc    & 42.9M   & 138.0G  & 40   & 52.8  & 71.0 & 34.2 & 57.3 & 70.0 & \\
                YOLOv8-S~\cite{yolov8}            & PAN      & 11.2M  & 28.6G   & 189  & 45.0  & 61.8  & 26.0 & 49.9 & 61.0 & \\
                YOLOv10-S~\cite{yolov10}          & PAN      & 8.1M   & 24.8G   & 173  & 46.2  & 63.0  & 26.9 & 51.1 & 63.7 & \\
                YOLOv12-S~\cite{yolov12}          & PAN      & 9.1M   & 19.4G   & 112  & 47.6  & 64.5  & 28.3 & 52.7 & 65.9 & \\
                \midrule
                RT-DETRv1 R18   & H-Enc + FINE  & 21.2M  & 62.9G   & 78   & 47.3 (+0.8) & 64.4  & 30.1 & 50.8 & 63.5 & \\
                RT-DETRv1 R50   & H-Enc + FINE  & 44.0M  & 139.2G  & 40   & 53.3 (+0.5) & 71.3  & 35.6 & 57.6 & 70.4 & \\
                YOLOv8-S        & PAN + FINE    & 12.3M  & 29.6G   & 177  & 45.8 (+0.8) & 62.7  & 27.0 & 51.2 & 62.1 & \\
                YOLOv10-S       & PAN + FINE    & 9.0M   & 25.7G   & 153  & 46.7 (+0.5) & 63.6  & 28.2 & 51.4 & 63.2 & \\
                YOLOv12-S       & PAN + FINE    & 10.5M  & 21.2G   & 106  & 48.2 (+0.6) & 65.1  & 30.6 & 53.3 & 65.2 & \\
                \midrule

                \rowcolor{gray!10}
                \multicolumn{11}{c}{Classic Object Detectors} \\
                \midrule
                Faster R-CNN R50~\cite{FasterR-CNN} & FPN           & 41.8M & 134.4G & -  & 37.0 & 58.5  & 21.1 & 40.3 & 48.2 & \\
                RetinaNet R50~\cite{RetinaNet}    & FPN           & 34.0M & 151.5G & -  & 36.4 & 55.7  & 19.1 & 40.0 & 48.9 & \\
                FCOS R50~\cite{FCOS}         & FPN           & 32.3M & 128.2G & -  & 39.2 & 58.2  & 22.1 & 42.4 & 51.3 & \\
                \midrule
                Faster R-CNN R50 & FPN + FINE  & 42.8M & 135.5G & -  & 39.1 (+2.1) & 60.8  & 23.3 & 42.6 & 50.4 & \\
                RetinaNet R50    & FPN + FINE  & 35.0M & 152.9G & -  & 37.3 (+0.9) & 56.8  & 20.6 & 40.7 & 49.4 & \\
                FCOS R50         & FPN + FINE  & 33.3M & 129.6G & -  & 39.8 (+0.6) & 58.7  & 23.3 & 43.1 & 51.2 & \\
            \bottomrule
            \end{tabular}
        }
    \end{center}
    \caption{
        \textbf{Performance of the proposed FINE module on various detectors} on COCO val2017.
        Models with `$^*$' mark yield slightly lower performance compared to the results reported in the original paper~\cite{RTDETRv1}, which may stem from differences in experimental setup or unreleased training details.
        PAN: PANet-style Fusion Network. H-Enc: Hybrid Encoder. 
    }
    \label{table:main_result}
\end{table*}

\begin{figure*}[!t]
\centering
\includegraphics[width=0.80\textwidth]{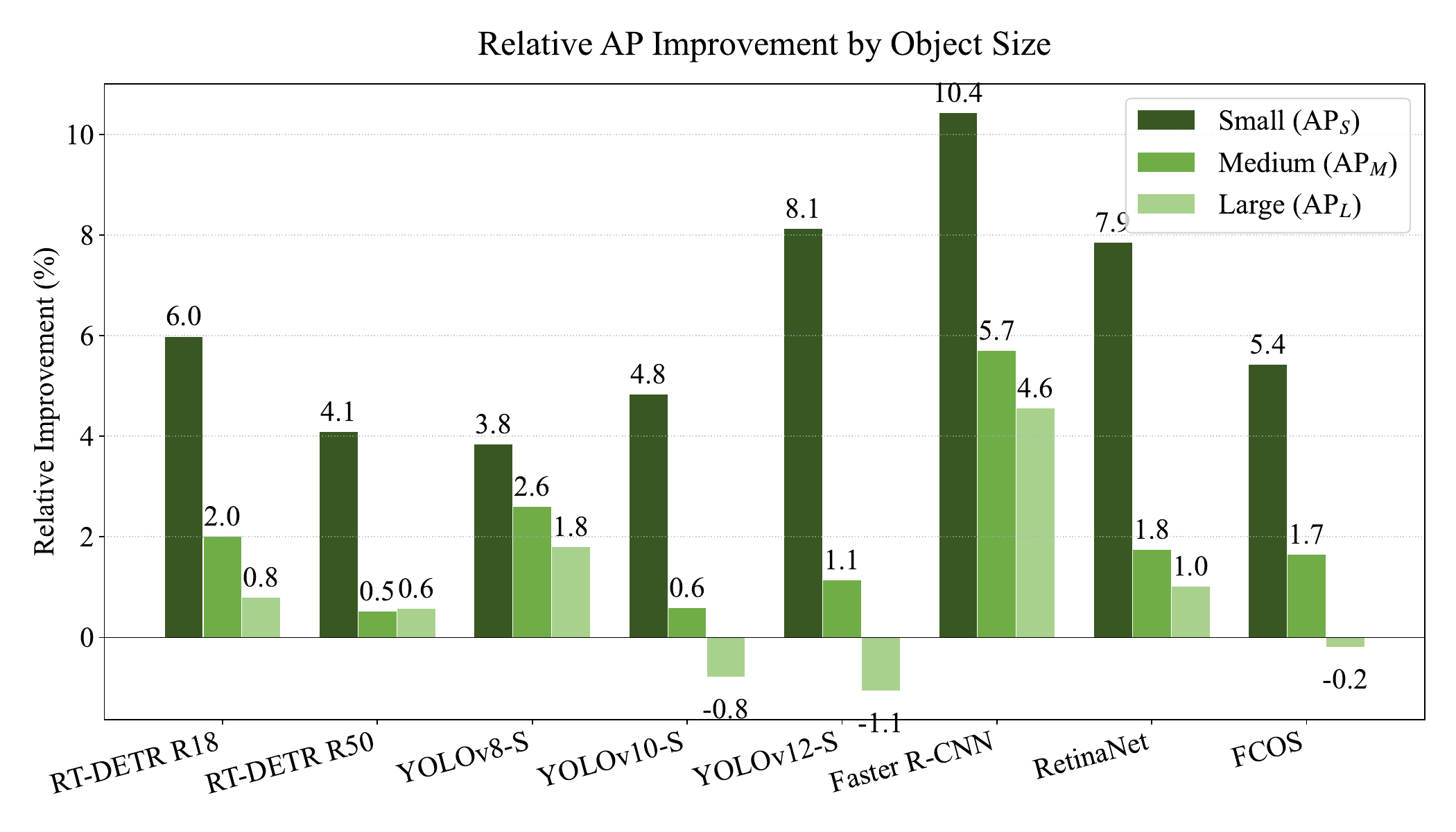} 
\vspace{-2mm}
\caption{
    \textbf{Relative AP improvement across different object sizes.} 
    The proposed FINE module achieves the most significant gains in small object detection.
}
\label{fig:relative_improvement}
\end{figure*}

\subsection{Main Results}

\subsubsection{Performance on Various Detectors}

To validate the generality of our approach, we integrate FINE into a diverse range of detectors, as summarized in Table~\ref{table:main_result}.
For CNN-based one-stage real-time detectors such as YOLOv8-S~\cite{yolov8}, FINE improves AP from 45.0 to 45.8 with negligible overhead (+1.0G FLOPs, +1.1M parameters).
We also evaluate FINE on RT-DETR-R50~\cite{RTDETRv1}, an end-to-end transformer-based detector with a hybrid encoder-decoder architecture. FINE increases AP from $52.8$ to $53.3$ while maintaining real-time throughput at $40$ FPS.
Lastly, on the attention-centric real-time detector YOLOv12-S~\cite{yolov12}, FINE yields a steady improvement from $47.6$ to $48.2$ AP.
These consistent results demonstrate that FINE addresses a structural limitation overlooked by existing detectors, providing reliable, architecture-agnostic improvements.

While FINE improves accuracy across most scales, its gains are particularly pronounced for small objects, as shown in Figure~\ref{fig:relative_improvement}. 
By mitigating semantic inconsistency in early fusion architectures like FPN~\cite{FPN}, FINE achieves substantial $\text{AP}_S$ improvements: $10.4\%$ for Faster R-CNN~\cite{FasterR-CNN}, $7.9\%$ for RetinaNet~\cite{RetinaNet}, and $5.4\%$ for FCOS~\cite{FCOS}. It also improves modern detectors with advanced fusion schemes, yielding $\text{AP}_S$ gains of $6.0\%$ on RT-DETR R18~\cite{RTDETRv1}, $4.8\%$ on YOLOv10-S~\cite{yolov10}, and $8.1\%$ on YOLOv12-S~\cite{yolov12}. These gains stem from enriching low-level features with high-level semantic context, which is particularly beneficial for small objects with limited semantic abstraction.
However, certain YOLO models show a marginal decrease in $\text{AP}_L$, which we attribute partly to the channel adaptation bottleneck in YOLO architectures (Appendix Section~\ref{appen_subsec:channel_unification}). Passing high-level features through this bottleneck to generate Keys and Values couples the optimization objectives of both scales within a compressed channel space. Although this may slightly compromise large-object representation, it remains a favorable trade-off given the substantial $\text{AP}_S$ gains.

\subsubsection{Comparison with Prior Cross-Scale Feature Alignment Methods}

We compare FINE against representative feature alignment methods on two distinct detector architectures: Faster R-CNN R50~\cite{FasterR-CNN} and RT-DETR R18~\cite{RTDETRv1}.
As shown in Table~\ref{table:VS_related}, FINE achieves the best accuracy-efficiency trade-off in both cases. 
On Faster R-CNN R50, FINE improves AP from 37.0 to 39.1 with only +1.0M parameters and +1.1G FLOPs. 
In contrast, MGC~\cite{A2-FPN} consumes +27.4G FLOPs for a smaller +1.3 AP gain, as its global context aggregation and redistribution across all pyramid levels introduce substantial computational overhead.
Furthermore, FINE achieves competitive accuracy while requiring substantially fewer resources than FaPN~\cite{FaPN} and AdaFPN~\cite{AdaFPN}, which add +9.2G and +25.0G FLOPs, respectively. 
A similar trend is observed on RT-DETR R18.
While several prior methods bring marginal or even negative changes, such as SNI~\cite{SNI} and AdaFPN~\cite{AdaFPN}, degrading the baseline by 0.2 and 0.4 AP, respectively, FINE achieves the largest improvement from 38.7 to 39.7 AP with minimal overhead (+1.0M parameters, +1.2G FLOPs). 
These results demonstrate that resolving semantic misalignment through lightweight cross-level attention is more efficient and robust than heavier architectural modifications or sampling-based strategies.

\begin{table*}[!ht]
    \begin{center}
        \resizebox{\textwidth}{!}{
            \begin{tabular}{lccc}
                \multicolumn{4}{c}{Faster R-CNN R50~\cite{FasterR-CNN}} \\
                \cmidrule(lr){1-4}
                Fusion Method & \#Params & FLOPs & $\Delta$ AP \\
                \midrule
                baseline      & 41.8M  & 134.4G  &  - \\
                $+$ SNI$^*$~\cite{SNI}                & 41.8M  & 134.4G  & 37.0 $\rightarrow$ 37.7 \\
                $+$ FaPN~\cite{FaPN}                  & 48.5M  & 143.6G  & 37.9 $\rightarrow$ 39.2 \\
                $+$ AdaFPN~\cite{AdaFPN}              & 45.6M  & 159.4G  & 37.8 $\rightarrow$ 39.0 \\
                $+$ $A^2$-FPN(MGC)$^*$~\cite{A2-FPN} & 44.4M  & 161.8G  & 37.0 $\rightarrow$ 38.3 \\
                \textbf{$+$ FINE (Ours)}  & \textbf{42.8M}  & \textbf{135.5G}  & \textbf{37.0 $\rightarrow$ 39.1} \\
                \bottomrule
            \end{tabular}
            
            \quad
            
            \begin{tabular}{lccc}
                \multicolumn{4}{c}{RT-DETR R18~\cite{RTDETRv1}} \\
                \cmidrule(lr){1-4}
                Fusion Method & \#Params & FLOPs & $\Delta$ AP \\
                \midrule
                baseline      & 20.2M  & 61.7G   & - \\
                $+$ SNI$^*$~\cite{SNI}                & 20.2M  & 61.7G   & 38.7 $\rightarrow$ 38.5 \\
                $+$ FaPN$^*$~\cite{FaPN}              & 21.9M  & 71.8G   & 38.7 $\rightarrow$ 39.0 \\
                $+$ AdaFPN$^*$~\cite{AdaFPN}          & 22.9M  & 82.8G   & 38.7 $\rightarrow$ 38.3 \\
                $+$ $A^2$-FPN(MGC)$^*$~\cite{A2-FPN} & 21.1M  & 68.7G   & 38.7 $\rightarrow$ 38.6 \\
                \textbf{$+$ FINE (Ours)}  & \textbf{21.2M}  & \textbf{62.9G}   & \textbf{38.7 $\rightarrow$ 39.7} \\
                \bottomrule
            \end{tabular}
        }
    \end{center}
    
    \caption{
        \textbf{Comparison with prior alignment methods.}
        Methods with `$^*$' are re-implemented and trained either according to the original paper or a Torchvision reference recipe~\cite{Torchvision}. 
        `MGC’ denotes the use of only the Multi-level Global Context module from $A^2$-FPN.
    }
    \label{table:VS_related}
\end{table*}

\begin{table*}[!th]
    \begin{center}
        \resizebox{0.96\textwidth}{!}{
            \renewcommand{\arraystretch}{1.2}
            \begin{tabular}{l|l|l|r|r|r|r|r|r}
                \toprule
                Model & Attention & Complexity & \#Params & FLOPs & Mem (MB) & FPS & Latency & AP \\
                \midrule
                RT-DETR R18 & -- & -- & 20.2M & 61.74G & 64.50 & 80 & 12.81ms & 38.7 \\
                \midrule
                
                \multirow{5}{*}{\textbf{+ FINE}} 
                & Vanilla~\cite{AttentionIsAllYouNeed} & $O(H_{\text{low}} W_{\text{low}} \times H_{\text{high}} W_{\text{high}})$ & \multirow{5}{*}{+1.0M} & +18.0G & +1223.48 & 57 & 17.74ms & 39.5 \\
                
                & Window~\cite{SwinT} & $O(H_{\text{low}} W_{\text{low}} M^2)$ & & +8.9G & +65.11 & 68 & 15.05ms & \textbf{39.7} \\
                
                & Area~\cite{yolov12} & $O(\frac{H_{\text{low}}W_{\text{low}} \times H_{\text{high}}W_{\text{high}}}{A})$ & & +10.0G & +65.67 & 69 & 14.71ms & 39.2 \\
                
                & Linear~\cite{SimA} & $O(H_{\text{low}} W_{\text{low}} + H_{\text{high}} W_{\text{high}})$ & & +7.0G & +18.42 & 64 & 16.05ms & 39.0 \\
                
                & \textbf{AATS (Ours)} & $O(\frac{H_{\text{low}} W_{\text{low}}}{r^2k^2} \times \frac{H_{\text{high}} W_{\text{high}}}{k^2})$ & & \textbf{+1.2G} & \textbf{+1.17} & \textbf{78} & \textbf{13.24ms} & \textbf{39.7} \\ 
                
                \bottomrule
            \end{tabular}
        }
    \end{center}
    \caption{
        \textbf{Performance of various attention strategies} integrated into the FINE module, evaluated on RT-DETR R18 ($1\times$ schedule).
        Window attention uses a window size of $7 \times 7$ ($M=7$)~\cite{SwinT}, and area attention divides feature maps into four regions ($A=4$)~\cite{yolov12}.
    }
    \label{table:attention_variants}
\end{table*}

\subsection{Ablation Study}
To assess the contribution of each component, we ablate four core designs of FINE on RT-DETR R18~\cite{RTDETRv1} and Faster R-CNN R50~\cite{FasterR-CNN} under a 1$\times$ schedule: the cross-level attention strategy, alignment-aware sampling ratio, number of attention heads, and modulation design. Additional ablations are also provided in Appendix Section~\ref{sec:appendix_additional_ablation}.


\subsubsection{Cross-level Attention Strategy}
\label{subsubsec:attention_strategy}
We evaluate several attention variants within FINE on RT-DETR R18. 
As reported in Table~\ref{table:attention_variants}, vanilla attention~\cite{AttentionIsAllYouNeed} incurs prohibitive quadratic overhead (+18.0G FLOPs, +1223.48 MB memory), reducing FPS from 80 to 57, while providing suboptimal gains (38.7 $\to$ 39.5 AP) due to unaligned cross-level ERFs. 
Local attention variants, including window~\cite{SwinT} and area~\cite{yolov12} attention, reduce computational cost but still introduce considerable overhead (+8.9G to +10.0G FLOPs, +65.11 to +65.67 MB memory) and restrict long-range dependencies. 
Linear attention~\cite{SimA} further reduces the overhead through kernel approximation, but remains costly when applied to high-resolution features (+7.0G FLOPs, +18.42 MB memory). 
Since these methods are primarily designed for intra-level interactions, they fail to address the underlying structural misalignment across levels.
In contrast, Alignment-Aware Token Sampling (AATS) explicitly aligns ERF scales before attention, achieving the best accuracy (39.7 AP) with negligible overhead (+1.2G FLOPs, +1.17 MB memory). 
Compared with vanilla attention, AATS reduces computational and memory costs by 93.3\% and 99.9\%, respectively, while preserving real-time speed (80 $\to$ 78 FPS).

\begin{figure*}[!t]
    \centering
    \begin{minipage}{0.44\textwidth}
        \centering
        \includegraphics[width=1.0\linewidth]{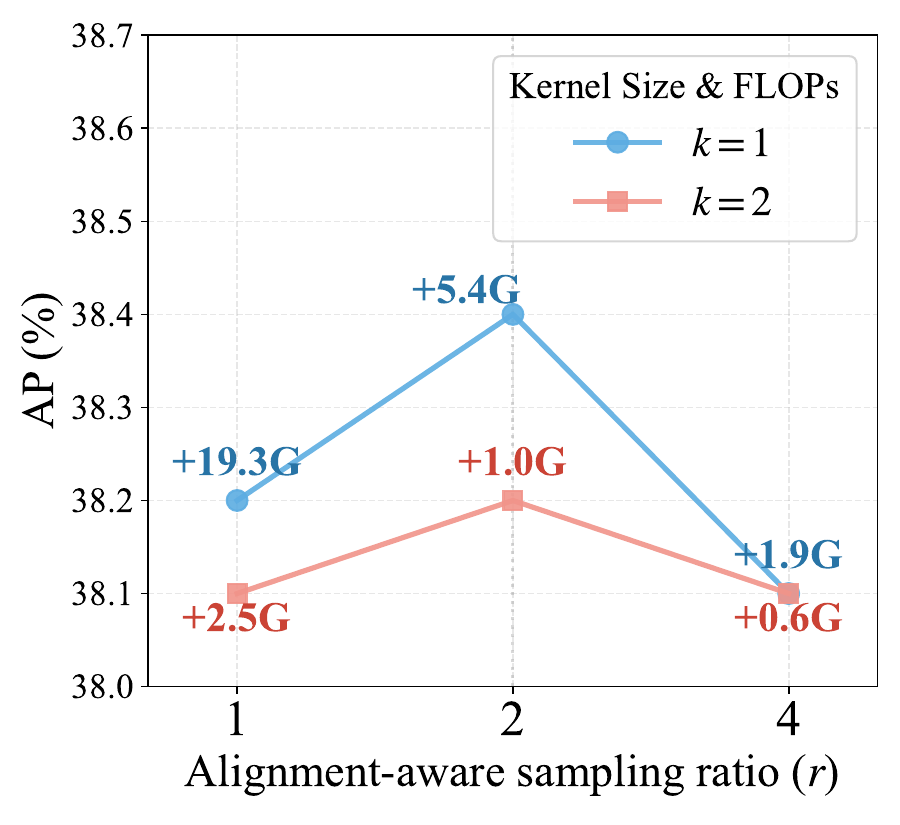} 
        \vspace{-3mm}
        \caption{
            \textbf{Ablation on AATS sampling ratio $r$.} 
            Aligning $r$ with the backbone's stride ($r=2$) provides the best performance-efficiency trade-off.
        }
        \label{fig:ablation_r}
    \end{minipage}
    \hfill
    \begin{minipage}{0.54\textwidth}
        \centering
        \includegraphics[width=\linewidth]{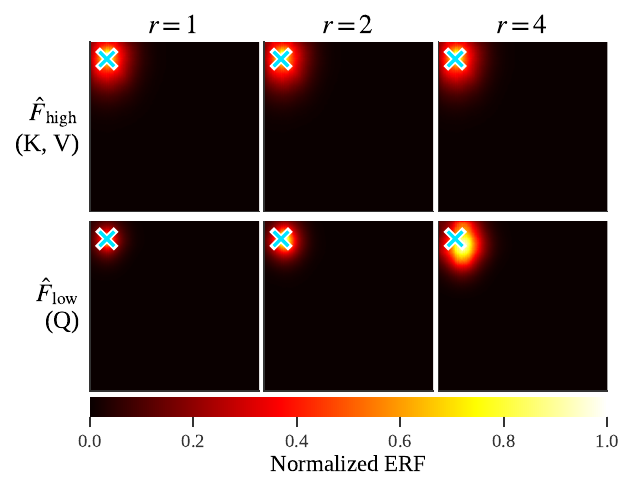} 
        \vspace{-3mm}
        \caption{
            \textbf{Visualization of ERFs across different sampling ratios $r$.} 
            The cyan `X' denotes the target token, selected at a normalized spatial coordinate of ($0.1, 0.1$) to ensure the exact same physical anchor point on the input image regardless of the feature map resolutions. 
        }
        \label{fig:ERF_heatmap_paired}
    \end{minipage}
\end{figure*}



\subsubsection{Alignment-Aware Sampling Ratio}
\label{subsubsec:align_aware_samplnig_ratio}
The core hyperparameter of AATS is the alignment-aware sampling ratio $r$, which
controls the degree to which the ERFs of $\hat{F}_\text{low}$ and $\hat{F}_\text{high}$
are synchronized.
Figure~\ref{fig:ablation_r} reports AP and FLOPs overhead on Faster R-CNN R50 for $r \in \{1, 2, 4\}$
and two kernel sizes $k \in \{1, 2\}$.
While $r = 4$ incurs the least overhead (+1.9G for $k=1$; +0.6G for $k=2$), it yields lower accuracy than $r = 2$, dropping from 38.4 to 38.1 AP for $k=1$, and from 38.2 to 38.1 AP for $k=2$.
Conversely, $r = 1$ introduces heavy computational costs (+19.3G for $k=1$; +2.5G for $k=2$) with no accuracy benefits (38.2 and 38.1 AP, respectively).
Setting $r = 2$ thus achieves the best accuracy-efficiency trade-off across both
kernel sizes.

Figure~\ref{fig:ERF_heatmap_paired} provides an ERF~\cite{ERF} visualization
that explains this result. 
For each cell, the ERF of a target token, marked
by the cyan `X' at the normalized coordinate $(0.1, 0.1)$, is computed by
back-propagating the token's response to the input image and rendered as a heatmap (brighter colors indicate stronger contribution). 
The normalized anchor keeps the physical reference consistent across resolutions, so comparing the two rows within each column reveals whether the ERFs are aligned at that $r$.
When $r = 1$, the sampled low-level tokens span a smaller physical footprint than the high-level tokens, causing a scale mismatch.
In contrast, $r = 4$ over-expands the low-level receptive field beyond the target region covered by the high-level tokens.
Only at $r=2$ do the ERFs of the two layers align closely, with the sampled
tokens covering comparable physical regions around the anchor point and
enabling reliable regional correspondences.
Together with the ERF-size analysis in Appendix Section~\ref{appendix_sec:ERF}, this provides both theoretical and empirical justification for choosing $r=2$.
Guided by these observations and prioritizing a lightweight design, we set $k = 1$ for the $l = 3$ fusion stage and $k = 2$ for the $l = 2$ fusion stage.

\subsubsection{Number of Attention Heads}
\label{subsubsec:number_of_attention_heads}

To investigate the multi-head mechanism within FINE, we ablate the number of attention heads $h$ on Faster R-CNN R50 with a fixed channel dimension $C = 256$.
This ablation aims to analyze the fundamental trade-off between maximizing subspace diversity for cross-level interactions and preserving sufficient per-head representational capacity.
As reported in Table~\ref{tab:head_ablation}, performance peaks at 38.2 AP when $h = 16$ or $h = 32$, where each head operates at a dimension of $d = 16$ or $d = 8$, respectively.
With fewer heads, such as $h = 1$ or $h = 4$, the limited number of shared subspaces restricts the diversity of cross-level interactions, yielding 37.9 AP and 38.1 AP, respectively.
Conversely, increasing $h$ beyond 32 reduces the per-head dimension below 8, which compresses the representational capacity of each subspace and degrades performance (e.g., dropping to 38.0 AP at $h = 128$).
Based on this analysis, we fix the per-head dimension to $d = 16$ as our default setting by adjusting $h$ accordingly.

\begin{table}[!t]
\begin{center}
\scalebox{0.80}{
    \setlength{\tabcolsep}{8pt}
    \begin{tabular}{c|c|ccccccc}
    \toprule
    \multirow{2}{*}{Metric} & Baseline & \multicolumn{7}{c}{Number of Heads ($h$) in FINE, given $C = 256$} \\
    \cline{3-9}
     & (w/o FINE) & 1 & 4 & 8 & \textbf{16} & \textbf{32} & 64 & 128 \\
    \midrule
    Per-Head Dim ($d=C/h$) & - & 256 & 64 & 32 & \textbf{16} & \textbf{8} & 4 & 2 \\
    \midrule
    AP & 37.0 & 37.9 & 38.1 & 38.1 & \textbf{38.2} & \textbf{38.2} & 38.1 & 38.0 \\
    \bottomrule
    \end{tabular}
}
\end{center}
\caption{
    \textbf{Ablation on the number of attention heads $h$.} 
    Evaluated on Faster R-CNN R50 with COCO val2017 (1 $\times$ schedule).
    With the total channel dimension fixed to $C = 256$, varying $h$ reveals the trade-off between subspace diversity and per-head capacity.
}
\label{tab:head_ablation}
\end{table}

\begin{figure*}[!t]
\centering
\includegraphics[width=1.0\textwidth]{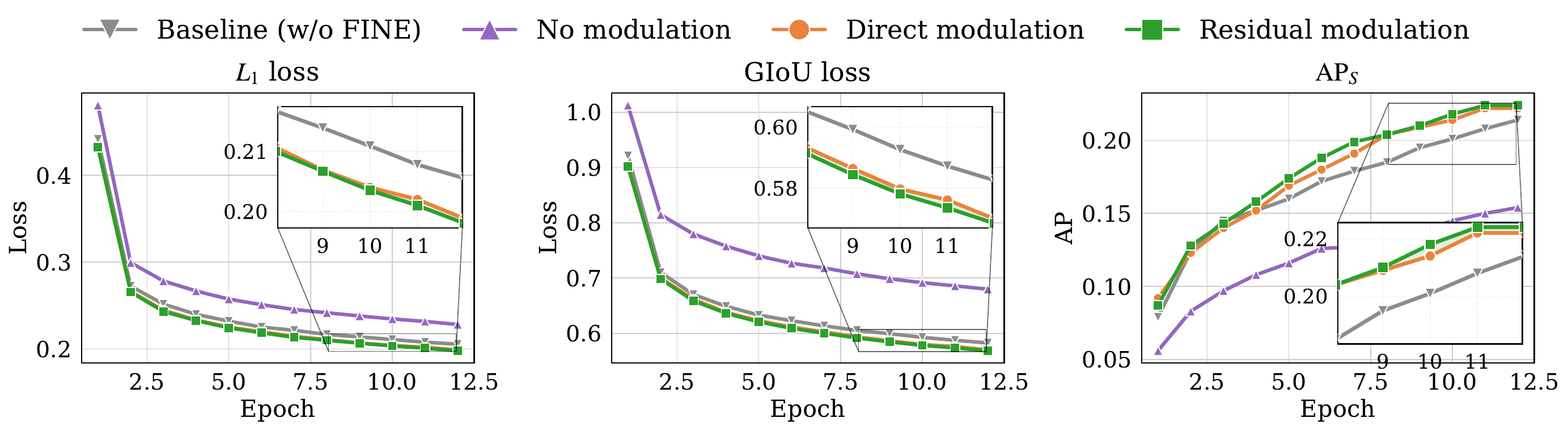} 
\vspace{-6mm}
\caption{
    \textbf{Ablation of the modulation strategy on localization losses and 
    small object detection.} 
    Training curves for RT-DETR R18 on COCO train2017 (1$\times$ schedule).
    No modulation underperforms even the FINE-free baseline on both metrics, 
    whereas repurposing the attention output as a modulation map (Direct and 
    Residual) consistently improves localization learning and $\text{AP}_S$.
    The Residual variant maintains marginally lower losses and higher 
    $\text{AP}_S$ than Direct modulation throughout training (zoomed insets).
}
\label{fig:residual_modulation_reg}
\end{figure*}

\subsubsection{Residual Modulation}
\label{subsubsec:modulation_startegy}

We ablate two design choices from Section~\ref{subsec:residual_modulation} on RT-DETR R18 by evaluating (i) the cross-level attention output as a directly refined feature versus a modulation map for $F_{\text{low}}$, and (ii) direct modulation via $F_{\text{low}} \odot M$ versus the residual formulation in Eq.~\ref{eq:residual_modulation}. 
Figure~\ref{fig:residual_modulation_reg} reports the $L_1$ and GIoU training losses alongside small object Average Precision ($\text{AP}_S$). 
Using the attention output directly as a refined feature (no modulation) substantially increases localization losses and reduces $\text{AP}_S$ from 21.4 (baseline) to 15.4, confirming that spatially coarse bottleneck signals cannot replace the high-resolution $F_{\text{low}}$. 
Conversely, repurposing the attention output as a modulation map reduces losses and improves $\text{AP}_S$ beyond the baseline, reaching 22.2 (direct) and 22.4 (residual). 
We attribute the localization and $\text{AP}_S$ advantages of the residual variant to its additive formulation. 
While direct modulation rescales $F_{\text{low}}$ by the coarse map $M$ and inevitably attenuates fine-grained spatial cues, the residual approach preserves the original high-resolution representation through additive refinement.
Preserving these sub-pixel cues is particularly important for small objects, where even minor losses of fine-grained spatial details can significantly affect object localization.
Further ablations regarding this residual design are detailed in Appendix Section~\ref{subsec:appendix_residual_modulation}.

\begin{figure*}[!t]
\centering
\includegraphics[width=0.9\textwidth]{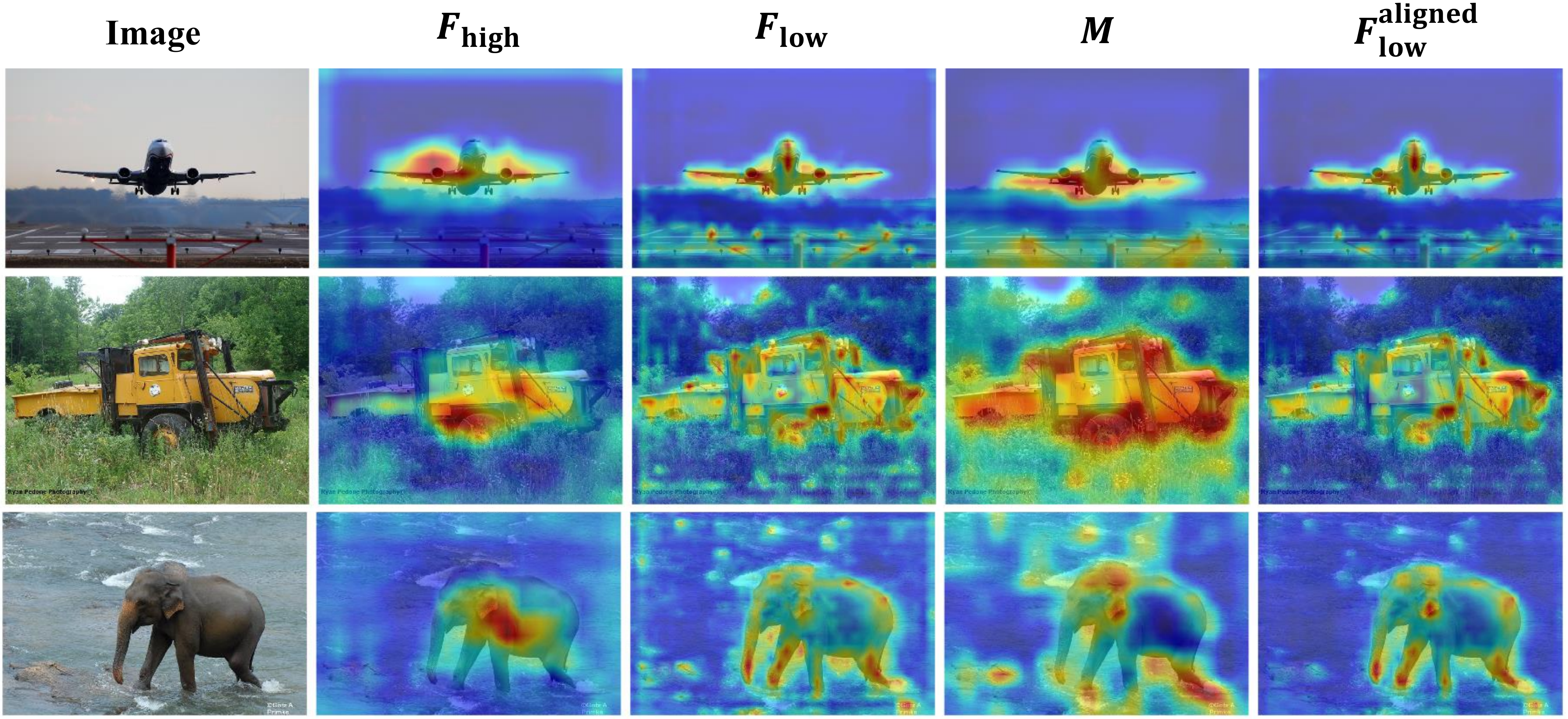} 
\caption{
    \textbf{Activation heatmaps} of high-level, low-level, learned modulation map, and aligned low-level features, visualized at the $l=3$ fusion stage on COCO.
}
\label{fig:activation_heatmap}
\end{figure*}

\begin{table}[!t]
    \begin{center}
        \resizebox{0.85\columnwidth}{!}{
            \begin{tabular}{lccccccc}
                \toprule
                               & \#Images & \#TPs ($\uparrow$) & \#FPs ($\downarrow$) & \#FNs ($\downarrow$) & Precision & Recall & F1-score \\ 
                \midrule
                FPN         & 5,000 & 23,833 & 21,557 & 12,502 & 0.525 & 0.656 & 0.583 \\
                FPN \textbf{+ FINE} & 5,000 & \textbf{24,186} & \textbf{19,417} & \textbf{12,149} & \textbf{0.555} & \textbf{0.666} & \textbf{0.605} \\ 
                \midrule
                Improvement & - & +1.48\% & -9.93\% & -2.82\% & +5.71\% & +1.52\% & +3.77\% \\ 
                \bottomrule
            \end{tabular}
        }
    \end{center}
    \caption{
        \textbf{Detection error analysis} of Mask R-CNN R50 on COCO val2017.
        Results are computed with an IoU threshold of 0.5 and confidence threshold of 0.5.
    }
    \label{tab:error_analysis}
\end{table}

\subsection{Analysis}
\label{subsec:exp_analysis}

To better understand how mitigating semantic inconsistency improves detection performance, we analyze activation heatmaps and detection errors using Mask R-CNN R50~\cite{MaskR-CNN} trained with a $2\times$ schedule on COCO.

\textbf{Activation Heatmap.} 
Activation heatmaps are visualized in Figure~\ref{fig:activation_heatmap}. 
Taking the first row as an example, $F_{\text{high}}$ concentrates activations on the airplane body, providing strong semantic cues but coarse spatial resolution, while $F_{\text{low}}$ captures fine-grained boundaries but also activates background regions.
Through cross-level interaction, the modulation map $M$ re-weights object-relevant regions relative to the background.
By applying residual modulation to preserve the spatial details of $F_{\text{low}}$, the resulting feature $F_{\text{low}}^{\text{aligned}}$ retains the sharp boundaries of the airplane while suppressing activations on background clutter. 

\begin{figure*}[!t]
\centering
\includegraphics[width=0.9\textwidth]{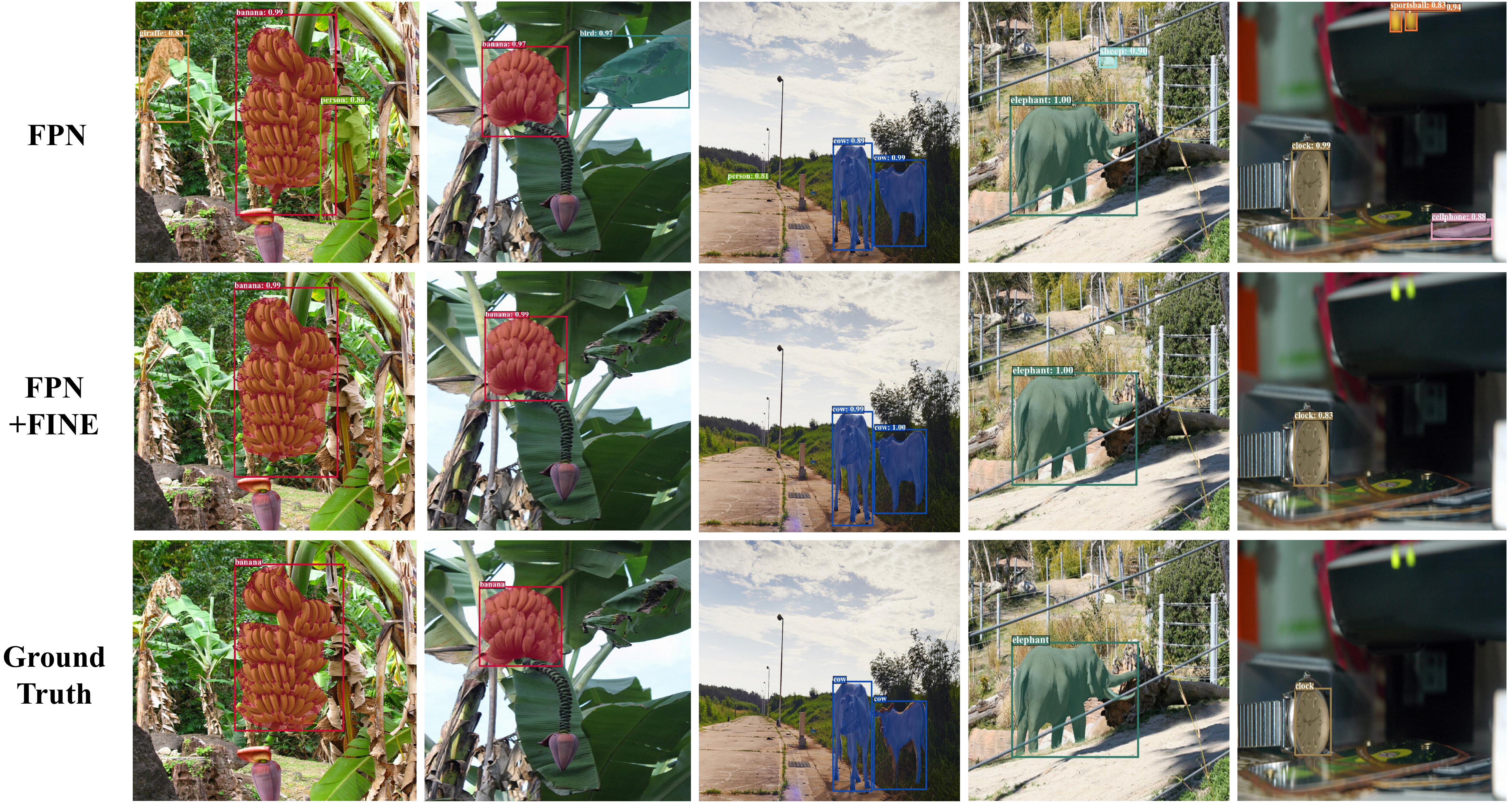} 
\caption{
    \textbf{False positive suppression comparing FPN~\cite{FPN} and FPN+FINE} on Mask R-CNN R50~\cite{MaskR-CNN} trained on COCO (confidence threshold: 0.8).
}
\label{fig:false_positives}
\end{figure*}

\textbf{Detection Error Analysis.} 
We further investigate whether the refined activations produced by FINE translate into fewer detection errors.
As shown in Table~\ref{tab:error_analysis}, FINE reduces False Positives (FPs) by 9.93\%, yielding a 5.71\% gain in precision, while a 2.82\% reduction in False Negatives (FNs) improves recall by 1.52\%. 
Importantly, this FP reduction does not suppress foreground predictions, as True Positives also increase ($23{,}833 \rightarrow 24{,}186$). Consequently, the overall F1-score improves by 3.77\%, indicating more reliable detection. 
As further illustrated in Figure~\ref{fig:false_positives}, FPN produces spurious detections on background clutter, such as complex textures and small structures, whereas FINE suppresses these high-confidence FPs while preserving correct predictions.
While FINE primarily improves precision in region proposal-based Mask R-CNN, likely by filtering dense proposals to suppress FPs, its benefit shifts toward improving object recall in the NMS-free, query-based RT-DETR (detailed in Appendix Section~\ref{subsec:error_analysis_RTDETR}).

\section{Conclusion}
\label{sec:conclusion}
We proposed a lightweight plug-and-play semantic alignment module, Feature Interaction NEtwork (FINE), to mitigate semantic inconsistency across hierarchical features by aligning low-level features with high-level contextual guidance. 
Through Alignment-Aware Token Sampling, FINE aligns effective receptive fields and establishes regional correspondences across pyramid levels, enabling bottleneck cross-level attention with minimal overhead. 
FINE consistently improves performance across diverse detectors, demonstrating semantic misalignment as a critical yet overlooked challenge in multi-scale feature fusion. 
While FINE currently operates within the conventional backbone--neck architecture by treating feature extraction and fusion as separate stages, future work could integrate cross-level interaction directly into feature extraction, unifying representation learning and multi-scale fusion.

\section*{Acknowledgements}
We thank the reviewers for their constructive comments and suggestions.
This work was supported by the National Research Foundation of Korea (NRF) grant funded by the Korea government (MSIT). (No. RS-2025-00522938)


\bibliography{egbib}


\clearpage


\appendix

\section{Implementation Details}
\label{appen_sec:implementation}

To facilitate reproducibility, we detail our training setup, the integration of FINE within standard FPN-based architectures, the specific channel adaptation steps for the YOLO series, and present our FPS measurement protocol with a component-level latency breakdown.

\subsection{Training Setup}
\label{appen_subsec:trainsetup}

For our experiments, real-time object detectors, reported in Table~\ref{table:main_result}, are trained using 4$\times$ NVIDIA GeForce RTX 3090 GPUs, while classic object detectors are trained on 2$\times$ NVIDIA RTX 6000 Ada GPUs. 
In all cases, we strictly adhere to the official training recipes and hyperparameters provided by the respective baseline implementations to ensure a fair and reproducible comparison.

\subsection{Architecture Integration and Pseudo-Code}
\label{appen_subsec:pseudo_code}

FINE is uniformly inserted before each top-down fusion stage (\eg, $l=3, 2$, as defined in Section~\ref{sec:pre}).
To provide concrete implementation details, Algorithm~\ref{alg:fine_pseudo} provides the PyTorch-like pseudo-code for integrating FINE within a standard top-down pathway.

\begin{algorithm}[h]
\caption{PyTorch-like Pseudo-Code for FINE in Top-Down Pathway}
\label{alg:fine_pseudo}
\begin{lstlisting}[language=Python, basicstyle=\ttfamily\scriptsize, keywordstyle=\color{blue}, commentstyle=\itshape\color{green!50!black}, frame=none]
import torch.nn as nn
import torch.nn.functional as F

class FeatureInteractionNEtwork(nn.Module):
    def __init__(self, dim, nhead, expansion):
        super().__init__()
        self.mha = nn.MultiheadAttention(dim, nhead)
        self.norm1 = nn.LayerNorm(dim)
        self.linear1 = nn.Linear(dim, dim * expansion)
        self.linear2 = nn.Linear(dim * expansion, dim)
        self.norm2 = nn.LayerNorm(dim)

    def forward(self, low_feat, high_feat, k_low, k_high):
        # 1. Alignment-Aware Token Sampling
        low_sampled = F.avg_pool2d(low_feat, kernel_size=k_low, stride=k_low)
        high_sampled = F.avg_pool2d(high_feat, kernel_size=k_high, stride=k_high)

        # 2. Bottleneck Multi-Head Cross-Level Attention
        q = low_sampled.flatten(2).transpose(1, 2)
        kv = high_sampled.flatten(2).transpose(1, 2)
        attn_out, _ = self.mha(query=q, key=kv, value=kv)

        # 3. Feed-Forward Network (FFN) & Residual & Norm
        attn_out = self.norm1(attn_out)
        attn_out = q + attn_out
        ffn_out = self.linear2(F.gelu(self.linear1(attn_out)))
        ffn_out = self.norm2(ffn_out)
        enc_out = attn_out + ffn_out

        # 4. Residual Spatial-Channel Modulation
        mod_map = enc_out.transpose(1, 2).reshape_as(low_sampled)
        mod_map = F.interpolate(mod_map, size=low_feat.shape[2:], mode='nearest')
        return low_feat + (mod_map * low_feat)


class TopDownPathway(nn.Module):
    def __init__(self, num_levels, dim, nhead, expansion, aats_ratio):
        super().__init__()
        self.r = aats_ratio
        self.fine = nn.ModuleList([
            FeatureInteractionNEtwork(dim, nhead, expansion) for _ in range(num_levels - 1)
        ])

    def forward(self, features):
        # features: list of multi-scale features [F2, F3, F4]
        out = [features[-1]] # Highest level feature
        L = len(features)
        for i in range(L - 2, -1, -1):
            low_feat = features[i]
            high_feat = out[0]

            depth = (L - 2) - i
            k_high = self.r ** depth
            k_low = k_high * self.r

            # ============ Feature Interaction NEtwork (FINE) ============
            low_feat = self.fine[depth](low_feat, high_feat, k_low, k_high)
            # ============================================================

            # Feature Fusion (e.g., standard FPN addition)
            upsampled_high = F.interpolate(high_feat, scale_factor=2.0, mode='nearest')
            fused_feat = low_feat + upsampled_high

            out.insert(0, fused_feat)
        return out
\end{lstlisting}
\end{algorithm}

\subsection{Channel Adaptation for YOLO Architectures}
\label{appen_subsec:channel_unification}

As described in Section~\ref{sec:pre}, most classic FPN-based detectors unify the channel dimensions of hierarchical features before fusion by applying $1\times1$ convolutions. 
This projects each feature map $\{S_2, S_3, S_4\}$ into a common embedding space with $C$ channels. 
This design simplifies subsequent fusion and attention operations, meaning no additional preprocessing is required for channel unification in such architectures.

In contrast, the YOLO series~\cite{yolov5, yolov6-v3.0, yolov8, yolov10, yolov12} retains the original backbone output channels $\{C_2, C_3, C_4\}$ without enforcing channel uniformity across scales. 
To enable cross-level attention within FINE for these architectures, the varying channel dimensions must be adapted. 
Specifically, we project both the low-level feature $S_l \in \mathbb{R}^{H_l \times W_l \times C_l}$ and the high-level feature $S_{l+1} \in \mathbb{R}^{H_{l+1} \times W_{l+1} \times C_{l+1}}$ to match the channel dimension of the low-level feature ($C_l$) using $1\times1$ convolutions followed by Batch Normalization:
\begin{equation}
    F_{\text{low}} = \text{BN}\big(\text{Conv}_{1\times1}(S_l)\big), \quad F_{\text{high}} = \text{BN}\big(\text{Conv}_{1\times1}(S_{l+1})\big)
\label{eq:appendix_channel}
\end{equation}
where $F_{\text{low}} \in \mathbb{R}^{H_l \times W_l \times C_l}$ and $F_{\text{high}} \in \mathbb{R}^{H_{l+1} \times W_{l+1} \times C_l}$ are the channel-aligned features passed to the FINE module. 
This adaptation step is introduced only when architecturally required, preserving FINE's plug-and-play generality across diverse detector families.

\subsection{FPS Measurement Protocol and Latency Breakdown}
\label{appen_subsec:latency_breakdown}

\noindent\textbf{FPS Measurement Protocol.}
For edge-device evaluations, we compile and execute the models on an NVIDIA Jetson Orin Nano using TensorRT~\cite{tensorrt}. 
Latency is profiled via the \texttt{trtexec} tool~\cite{tensorrt_benchmarking} at FP16 precision with a batch size of $1$. 
To ensure stable operating conditions and prevent thermal throttling, the reported FPS metrics are averaged over 1,000 timed iterations following a 50-iteration warm-up. 
Following the evaluation protocol of RT-DETR~\cite{RTDETRv1}, Table~\ref{table:main_result} reports FPS exclusively for real-time detectors, although classic detectors exhibit similarly minimal computational overhead. 
The identical FPS reported for RT-DETR R50 in Table~\ref{table:main_result} is purely a rounding artifact due to FINE's minimal overhead (shifting only from $39.92$ to $39.65$ FPS, both rounding to $40$). 
For precise comparisons, Table~\ref{tab:exact_fps} provides the exact unrounded measurements.

\noindent\textbf{Per-component Latency Breakdown.} 
Table~\ref{tab:fine_latency_breakdown} details the layer-wise latency of the FINE module to analyze its computational distribution. 
Notably, the spatial reduction achieved by Alignment-Aware Token Sampling effectively bounds the cross-attention overhead to only $28.2\%$. 
Because the attention mechanism is significantly lightened, the largest computational contributor shifts to Residual Modulation ($30.9\%$), which inherently incurs higher latency by operating on the restored high-resolution feature maps. 
It is worth noting that while the sum of these individually profiled components totals $0.492$ ms due to measurement synchronization overhead, concurrent GPU execution in practice limits the actual end-to-end latency increase of FINE to a mere $0.17$ ms (e.g., from $25.05$ ms to $25.22$ ms on RT-DETR R50).

\begin{table*}[t]
    \centering
    \begin{minipage}[t]{0.48\textwidth}
        \centering
        \scalebox{0.85}{
        \begin{tabular}{lrr}
        \toprule
        Model & w/o FINE & w/ FINE \\
        \midrule
        RT-DETRv1 R18  & 80.99 & 77.93 \\
        RT-DETRv1 R50  & 39.92 & 39.65 \\
        YOLOv8-S       & 189.37 & 177.07 \\
        YOLOv10-S      & 172.99 & 152.58 \\
        YOLOv12-S      & 111.92 & 105.73 \\
        \bottomrule
        \end{tabular}
        }
        \vspace{2mm}
        \caption{
            \textbf{FPS without integer rounding} for real-time detectors reported in Table~\ref{table:main_result}.
        }
        \label{tab:exact_fps}
    \end{minipage}
    \hfill
    \begin{minipage}[t]{0.48\textwidth}
        \centering
        \scalebox{0.85}{
        \begin{tabular}{lrr}
        \toprule
        Component & Latency (ms) & Proportion \\
        \midrule
        AATS (Down)          & 0.077 & 15.6\% \\
        Cross-Attention  & 0.139 & 28.2\% \\
        FFN              & 0.045 &  9.2\% \\
        AATS (Up)         & 0.079 & 16.1\% \\
        Residual Modulation       & 0.152 & 30.9\% \\
        \midrule
        Total   & 0.492 & 100\% \\
        \bottomrule
        \end{tabular}
        }
        \vspace{-2mm}
        \caption{
            \textbf{Per-component latency of the FINE module,} profiled on RT-DETR R50 with TensorRT FP16 on NVIDIA Jetson Orin Nano.
        }
        \label{tab:fine_latency_breakdown}
    \end{minipage}
\end{table*}

\section{Multi-Seed Evaluation on RT-DETR R18}
\label{appendix_sec:multi_seed}

To demonstrate that the performance gains introduced by FINE are statistically significant and robust against stochastic optimization, we conduct a multi-seed evaluation on RT-DETR R18~\cite{RTDETRv1} (6$\times$, full schedule). 
Table~\ref{tab:rtdetr_r18_seed_variance} reports the COCO val2017 detection accuracy across 8 independent seeds. 
Notably, because the official implementation of the baseline RT-DETR defaults to seed 0, we adopt the exact same initialization for the main results reported in Table~\ref{table:main_result} to ensure a fair comparison.
The overall accuracy remains highly stable, achieving an average AP of $47.23$ with a standard deviation of $\pm0.04$. 
Notably, FINE consistently provides larger relative gains for smaller objects: AP$_S$ improves from $28.4$ to $29.75$ (+4.75\%), compared with +1.73\% for AP$_M$ and +0.78\% for AP$_L$. 
Although AP$_S$ exhibits relatively higher variance ($\pm0.49$), its performance consistently exceeds the baseline across all 8 seeds, demonstrating that the stronger gains on small objects are not driven by a particular random initialization.

\begin{table}[h]
\centering
\resizebox{0.95\linewidth}{!}{
\begin{tabular}{l | cccccc}
\toprule
Model / Seed & AP & AP$_{50}$ & AP$_{75}$ & AP$_{S}$ & AP$_{M}$ & AP$_{L}$ \\
\midrule
RT-DETR R18~\cite{RTDETRv1} & 46.5 & 63.8 & 50.4 & 28.4 & 49.8 & 63.0 \\
\midrule
+ FINE (Seed 0) & 47.3 & 64.4 & 51.2 & 30.1 & 50.8 & 63.5 \\
+ FINE (Seed 1)       & 47.2 & 64.2 & 51.2 & 30.3 & 50.8 & 63.3 \\
+ FINE (Seed 2)       & 47.3 & 64.3 & 51.3 & 29.2 & 50.6 & 63.8 \\
+ FINE (Seed 3)       & 47.2 & 64.0 & 51.2 & 29.4 & 50.7 & 63.3 \\
+ FINE (Seed 4)       & 47.2 & 64.2 & 51.0 & 30.6 & 50.7 & 63.6 \\
+ FINE (Seed 5)       & 47.2 & 64.2 & 51.1 & 29.3 & 50.7 & 63.7 \\
+ FINE (Seed 6)       & 47.2 & 64.2 & 51.3 & 29.7 & 50.7 & 63.3 \\
+ FINE (Seed 7)       & 47.2 & 64.1 & 51.1 & 29.4 & 50.3 & 63.4 \\
\midrule
+ FINE (Mean\,$\pm$\,Std)
  & $47.23_{\pm0.04}$
  & $64.20_{\pm0.11}$
  & $51.17_{\pm0.10}$
  & $29.75_{\pm0.49}$
  & $50.66_{\pm0.15}$
  & $63.49_{\pm0.18}$ \\
\bottomrule
\end{tabular}
}
\vspace{2mm}
\caption{
    \textbf{Per-seed detection accuracy of RT-DETR-R18 with FINE.}
    Evaluated over 8 independent seeds on COCO val2017. 
}
\label{tab:rtdetr_r18_seed_variance}
\end{table}

\section{Applicability to Dense Prediction Tasks}
\label{appen_subsec:other_dense}

Although FINE is primarily evaluated for object detection, its core principle of enhancing semantic consistency across hierarchical features is also applicable to pixel-level dense prediction tasks. 
To demonstrate this generality, we integrate FINE into representative frameworks for instance, semantic, and panoptic segmentation. 
For instance segmentation, we follow the Torchvision training recipe~\cite{Torchvision}, while semantic and panoptic segmentation experiments adopt the $1\times$ training schedule from OpenMMLab~\cite{mmsegmentation}, using input resolutions of $640 \times 640$ for COCO~\cite{COCO} and $512 \times 1024$ for Cityscapes~\cite{cityscapes}.

As shown in Table~\ref{table:dense}, FINE consistently improves performance across all three tasks with negligible computational overhead. 
For instance segmentation on COCO, adding FINE to Mask R-CNN R50~\cite{MaskR-CNN} improves box AP and mask AP by $+1.8$ and $+0.9$, respectively. 
For semantic segmentation on Cityscapes, incorporating FINE into Semantic FPN~\cite{semantic_panoptic_fpn} yields a $+1.9$ mIoU improvement. 
For panoptic segmentation on COCO, Panoptic FPN with FINE achieves $+0.7$ gains in PQ (Panoptic Quality), SQ (Segmentation Quality), and RQ (Recognition Quality), following the evaluation metrics of~\cite{panoptic_metric}.

These results demonstrate that although FINE extracts region-level semantic context due to its bottleneck architecture, the resulting feature representations are by no means coarse. 
By using this regional context to modulate high-resolution low-level features via a residual connection, FINE effectively broadcasts strong categorical cues to individual pixels while preserving fine-grained spatial details.

\begin{table}[!t]
    \begin{center}
        \scalebox{0.9}{
        \begin{tabular}{l | c | c  r | l l l}
            \multicolumn{7}{l}{Instance segmentation on COCO~\cite{COCO} dataset} \\
            \toprule
            Model & FINE & \#Params & FLOPs & Box AP & Mask AP & \\
            \midrule
            Mask R-CNN R50~\cite{MaskR-CNN}         & -- & 44.4M & 134.4G & 37.9                  & 34.6                  & \\
            Mask R-CNN R50         & \checkmark & 45.5M & 135.5G & 39.7 (+1.8) & 35.5 (+0.9) & \\
            \midrule
            \multicolumn{7}{c}{} \\[-0.8em]
            
            \multicolumn{7}{l}{Semantic segmentation on Cityscapes~\cite{cityscapes} dataset} \\
            \toprule
            Model & FINE & \#Params & FLOPs & mIoU & & \\
            \midrule
            Semantic FPN~\cite{semantic_panoptic_fpn}           & -- & 28.5M & 90.9G & 74.5                   & & \\
            Semantic FPN           & \checkmark & 30.1M & 91.7G & 76.4 (+1.9)  & & \\
            \midrule
            \multicolumn{7}{c}{} \\[-0.8em]
            
            \multicolumn{7}{l}{Panoptic segmentation on COCO dataset} \\
            \toprule
            Model & FINE & \#Params & FLOPs & PQ & SQ & RQ \\
            \midrule
            Panoptic FPN~\cite{semantic_panoptic_fpn}           & -- & 46.0M & 156.7G & 40.2                 & 77.8                 & 49.3 \\
            Panoptic FPN           & \checkmark & 47.6M & 157.3G & 40.9 (+0.7) & 78.5 (+0.7) & 50.0 (+0.7) \\
            \bottomrule
        \end{tabular}
        }
    \end{center}
    \caption{
        \textbf{Generality of FINE across dense prediction tasks.}
        Beyond object detection, FINE consistently enhances performance in instance, semantic, and panoptic segmentation with minimal computational overhead.
    }
    \label{table:dense}    
\end{table}

\section{Performance on VisDrone Dataset}
\label{appendix_sec:VisDrone}

\begin{table}[!t]
\centering
\setlength{\tabcolsep}{6pt} 
\begin{center}
    \scalebox{0.80}{
        \begin{tabular}{l|l c l}
            \toprule
            Model & Fusion Method & FLOPs & AP$_{50}$ \\
            \midrule
            YOLOv5-S~\cite{yolov5} & PAN & 37.5G & 45.9 \\
            & PAN \textbf{+ FINE} & 40.8G & \textbf{46.9 (+1.0)} \\
            \midrule
            YOLOv6-S v3.0~\cite{yolov6-v3.0} & PAN & 72.2G & 44.8 \\
            & PAN \textbf{+ FINE} & 74.5G & \textbf{45.6 (+0.8)} \\
            \midrule
            YOLOv8-S~\cite{yolov8} & PAN & 44.7G & 46.3 \\
            & PAN \textbf{+ FINE} & 49.2G & \textbf{47.2 (+0.9)} \\
            \bottomrule
        \end{tabular}
    }
\end{center}
\caption{
    \textbf{Performance on VisDrone-DET2019.} 
    All models are evaluated with an input resolution of $800 \times 800$.
}
\label{table:visdrone_800}
\end{table}

The VisDrone-DET2019 dataset~\cite{visdrone} is a challenging benchmark characterized by crowded scenes, complex backgrounds, and a large proportion of small objects captured from aerial viewpoints. 
To evaluate the robustness of FINE under these demanding conditions, we integrate it into several representative real-time detectors~\cite{yolov5, yolov6-v3.0, yolov8}. 
As summarized in Table~\ref{table:visdrone_800}, integrating FINE consistently improves $\text{AP}_{50}$ across all evaluated detectors, yielding gains of up to +1.0.

Notably, these YOLO baselines are primarily designed for generic object detection and real-time efficiency rather than for specialized small-object detection. 
Nevertheless, FINE consistently improves detection accuracy with marginal computational overhead, demonstrating that selectively modulating low-level features under high-level semantic guidance effectively benefits target localization in small-object-dominated scenarios.

\section{Effective Receptive Field Analysis}
\label{appendix_sec:ERF}

Following the theoretical framework established in~\cite{ERF}, we define the effective receptive field (ERF) of a feature activation $y_{i,j}$ at spatial location $(i,j)$ by measuring its gradient-based influence with respect to all pixels on the input plane. 
Formally, for any input pixel $x_{u,v}$ at spatial coordinates $(u,v)$, the influence is defined as
\begin{equation}
g_{u,v} = \left|\frac{\partial y_{i,j}}{\partial x_{u,v}}\right|.
\end{equation}

The ERF is characterized by the spatial distribution of $g_{u,v}$ across all input coordinates $(u,v)$.
Under a cascade of convolutional layers, the central limit theorem implies that this distribution asymptotically approaches a 2D Gaussian:
\begin{equation}
g_{u,v} \propto 
\exp\left(
-\frac{\|(u,v) - \boldsymbol{\mu}\|^2}{2\sigma^2}
\right),
\end{equation}
where $\boldsymbol{\mu}=(\mu_u,\mu_v)$ denotes the spatial center of the effective receptive field, and the standard deviation $\sigma$ is taken as \textit{the effective receptive field (ERF) size}, which approximates its spatial radius.

\begin{figure*}[!t]
\centering
\includegraphics[width=0.5\textwidth]{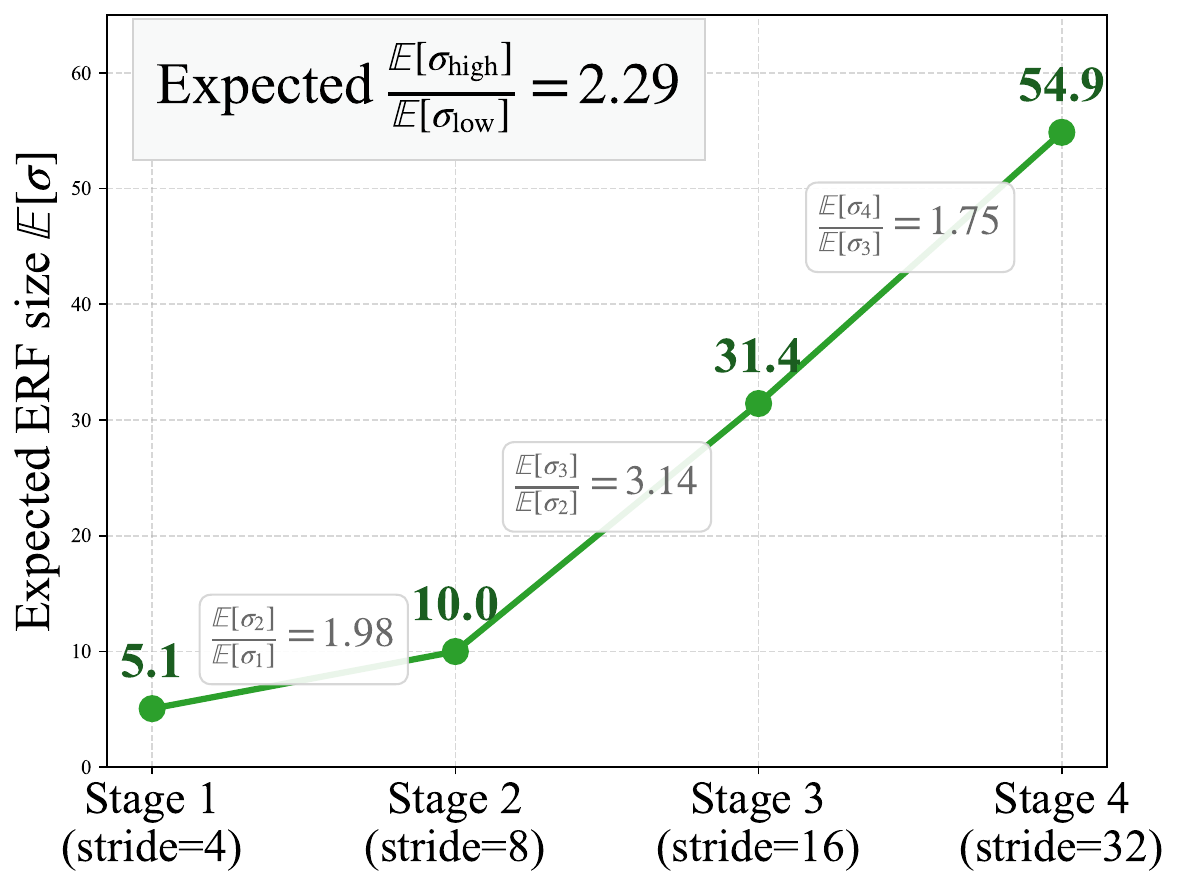} 
\caption{
    \textbf{Empirical analysis of effective receptive fields (ERFs)} in ResNet-50~\cite{ResNet}. 
    Evaluated over the COCO val2017, the average expected ERF expansion ratio across adjacent stages closely approximates the architectural stride ratio.
}
\label{fig:ERF}
\end{figure*}

To empirically validate the relationship between ERF size~\cite{ERF} and network downsampling, 
we conduct an ERF analysis on ResNet-50~\cite{ResNet} over all 5,000 images of the COCO val2017 split.
For each image, we compute $\sigma$ at the central spatial location of the 
final feature map within each backbone stage and average these values across the dataset 
to obtain $\mathbb{E}[\sigma]$ per stage.
As illustrated in Figure~\ref{fig:ERF}, the ratio of expected ERF sizes between 
adjacent stages closely approximates the architectural stride ratio:
\begin{equation}
    \frac{\mathbb{E}[\sigma_{\text{high}}]}{\mathbb{E}[\sigma_{\text{low}}]} \approx s,
\end{equation}
where $s$ denotes the stride ratio between adjacent stages.
Averaged across all adjacent stage pairs (Stage 1--2, 2--3, and 3--4), this 
ratio is $2.29$. This closely matches the theoretical stride ratio of $s=2$, 
providing an empirical justification for setting the alignment-aware 
sampling ratio to $r=2$ in our Alignment-Aware Token Sampling strategy.

\section{Additional Ablation Study}
\label{sec:appendix_additional_ablation}

\subsection{Downsampling Operator in Alignment-Aware Token Sampling}
\label{appen_subsec:downsampling}

We present an ablation study on the downsampling operator $\text{Down}(\cdot)$ used within our Alignment-Aware Token Sampling (AATS) (Eq.~\ref{eq:AATS}), with results summarized in Table~\ref{tab:downsampling_ablation}.
Although a learnable operator such as depthwise separable convolution (DWConv)~\cite{DWConv} could capture richer local statistics, 
parameter-free pooling operators (MaxPool and AvgPool) achieve comparable performance while introducing no additional parameters or computational overhead. 
This suggests that the primary role of $\text{Down}(\cdot)$ in AATS is not complex feature transformation, but rather the alignment of effective receptive fields across feature levels. 
Consequently, a simple pooling operator is sufficient to provide the required spatial aggregation for ERF alignment.
Based on these results, we adopt average pooling as the default implementation of AATS due to its simplicity, parameter-free design, and smoother preservation of regional feature distributions compared to the discrete selection behavior of max pooling.

\begin{table}[htbp]
    \begin{center}
        \begin{tabular}{l | c  c  c}  
            \toprule
            $\text{Down}(\cdot)$ & MACs & \#Params & AP \\
            \midrule
            DWConv                & 2.46M  & +6.14K  & 38.1 \\
            MaxPool               & --     & --      & 38.2 \\
            AvgPool               & --     & --      & 38.2 \\
            \bottomrule
        \end{tabular}
    \end{center}
    \caption{
        \textbf{Ablation on different downsampling operators in Alignment-Aware Token Sampling.}
        Results are evaluated using Faster R-CNN R50~\cite{FasterR-CNN} on COCO val2017 (1$\times$ training schedule).
    }
    \label{tab:downsampling_ablation}
\end{table}

\subsection{Impact of ERF Alignment and Sampling Strategies}

\begin{figure*}[!t]
\centering
\includegraphics[width=0.85\textwidth]{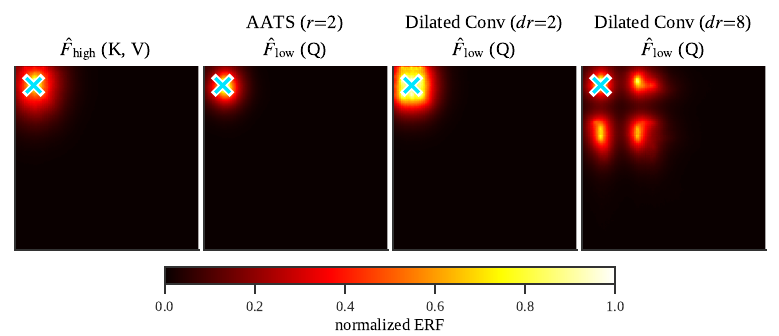} 
\caption{
    \textbf{Visualization of ERFs across different sampling strategies.} 
    The cyan `X' denotes the target token, selected at a normalized spatial coordinate of ($0.1, 0.1$) to ensure the exact same physical anchor point on the input image regardless of the feature map resolutions. 
    Compared to the spatial mismatches observed in dilated convolutions (where $dr$ denotes the dilation rate), AATS structurally aligns the ERFs across different levels.
}
\label{fig:ERF_adj_nonadj}
\end{figure*}

To validate that the performance gains of the FINE module stem from ERF alignment rather than simple resolution reduction, we further ablate our Alignment-Aware Token Sampling (AATS). 
We compare AATS with a dilated convolution baseline that achieves the same spatial resolution reduction but lacks structural ERF alignment. 
As shown in Table~\ref{table:appendix_sampling_strategy}, the dilated convolution variant underperforms our ERF-aware AATS strategy ($39.3$ vs.\ $39.7$ AP), despite introducing more parameters (22.8M vs.\ 21.2M) and computational overhead (64.2G vs.\ 62.9G FLOPs).
Supported by the ERF visualizations in Figure~\ref{fig:ERF_adj_nonadj}, this result indicates that merely reducing the spatial scale is insufficient. 
Structurally establishing cross-level correspondence is crucial for effective cross-attention between hierarchical features.

We also investigate the generality of our ERF-alignment principle by applying AATS to fuse non-adjacent pyramid levels (e.g., $F_2 \leftrightarrow F_4$). 
Following our design principle, the sampling ratio $r$ must match the stride ratio between the fused levels to guarantee ERF alignment. 
We set $r=4$ to account for the larger stride gap in this setup. 
Table~\ref{table:appendix_sampling_strategy} shows that ERF-aligned sampling maintains a highly competitive accuracy of $39.5$ AP even with non-adjacent fusion. 
This suggests that our formulation serves as a generalized principle to bridge semantic and scale gaps across diverse feature pyramid hierarchies, rather than a heuristic limited to adjacent layers.

\begin{table}[!t]
    \begin{center}
        \scalebox{0.9}{
        \begin{tabular}{l c c c c c}
            \toprule
            Sampling & ERF-aware & Fusion & Params & FLOPs & AP$_{50:95}$ \\
            \midrule
            Dilated Conv      & $\times$   & Adj.     & 22.8M & 64.2G & 39.3 (+0.6) \\
            \midrule
            AvgPool ($r{=}2$) & \checkmark & Adj.     & 21.2M & 62.9G & 39.7 (+1.0) \\
            AvgPool ($r{=}4$) & \checkmark & Non-Adj. & 21.2M & 62.9G & 39.5 (+0.8) \\
            \bottomrule
        \end{tabular}
        }
    \end{center}
    \caption{
        \textbf{Ablation of alignment-aware token sampling and fusion strategies} on RT-DETR R18 under a 1$\times$ schedule.
        Adj.: $F_2 \leftrightarrow F_3$; Non-Adj.: $F_2 \leftrightarrow F_4$.
        The performance gain is measured over the baseline without FINE.
        For Dilated Conv, the dilation rate is set to 2.
    }
    \label{table:appendix_sampling_strategy}
\end{table}

\subsection{Residual Modulation}
\label{subsec:appendix_residual_modulation}

As discussed in Section~\ref{subsubsec:modulation_startegy} and illustrated in Figure~\ref{fig:residual_modulation_reg}, repurposing the cross-level attention output as a modulation map instead of directly replacing features is crucial to avoid discarding high-resolution spatial details.
Furthermore, we introduce residual modulation as a conservative design to preserve fine-grained features from the original low-level representations. 
While this residual path does not significantly alter overall accuracy, it acts as a structural safeguard against the degradation of sub-pixel localization cues typically caused by spatially coarse modulation maps.
To provide a more robust comparison between direct and residual modulation, Table~\ref{tab:ablation_resmod_seed} reports their performance averaged over three random seeds. 
Both strategies achieve identical overall accuracy ($48.97$ AP). 
However, the scale-specific metrics reveal a distinct trade-off: residual modulation yields a higher AP$_S$ ($30.47$ vs.\ $29.93$) at the expense of a marginal reduction in AP$_L$ ($66.60$ vs.\ $66.97$). 
This observation confirms that the residual formulation functions as intended, prioritizing the preservation of fine-grained structural cues that are critical for localizing small objects. 

\begin{table}[!t]
    \begin{center}
        \scalebox{0.9}{
        \begin{tabular}{lcccc}
            \toprule
            RT-DETR R50 (1$\times$) & AP$_{50:95}$ & AP$_S$ & AP$_M$ & AP$_L$ \\
            \midrule
            direct   & 48.97 & 29.93 & 53.40 & \textbf{66.97} \\
            residual & 48.97 & \textbf{30.47} & \textbf{53.53} & 66.60 \\
            \bottomrule
        \end{tabular}
        }
    \end{center}
    \caption{
        \textbf{Performance comparison of direct and residual modulation.} 
        Averaged over three random seeds. 
        While the overall accuracy (AP$_{50:95}$) remains identical across both settings, the residual formulation effectively preserves high-resolution spatial details, leading to improvements in small object detection (AP$_S$).
    }
    \label{tab:ablation_resmod_seed}
\end{table}

\begin{figure*}[!t]
\centering
\includegraphics[width=1.0\textwidth]{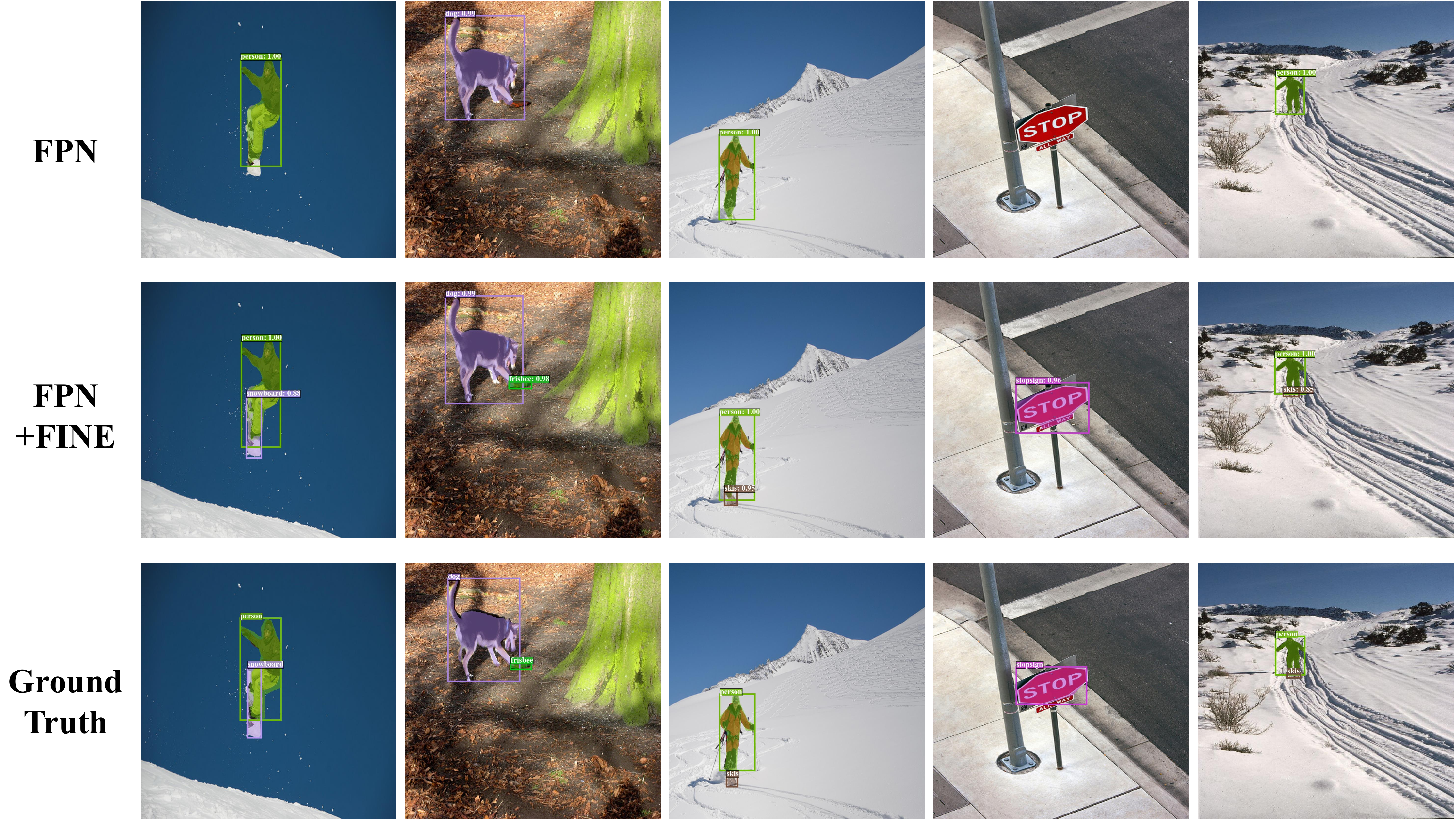} 
\caption{
    \textbf{False negatives reduction comparing FPN~\cite{FPN} and FPN+FINE} on Mask R-CNN R50~\cite{MaskR-CNN} trained on COCO (confidence threshold: 0.8).
}
\label{fig:appendix_FNs}
\end{figure*}

\begin{figure*}[!t]
\centering
\includegraphics[width=1.0\textwidth]{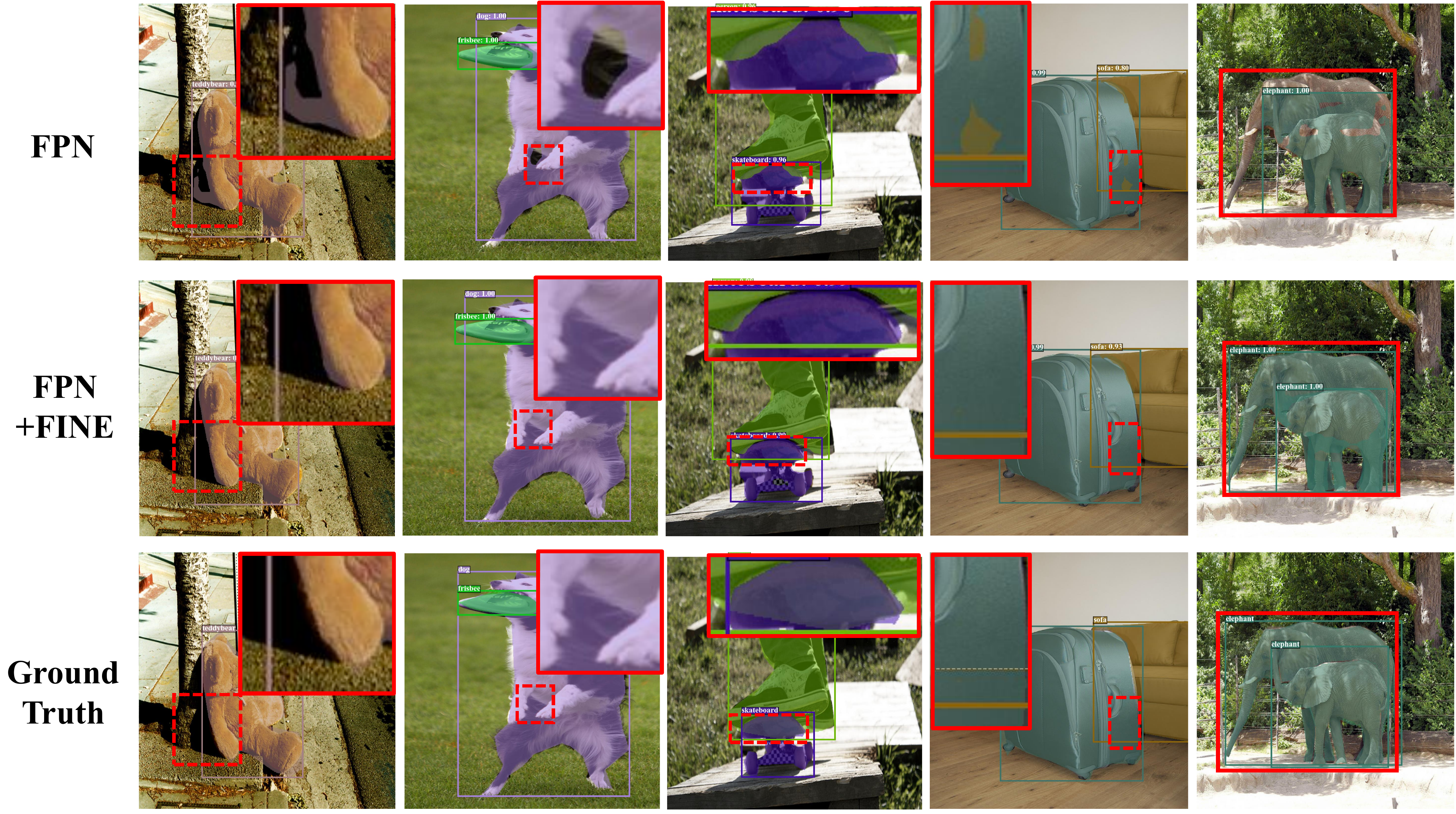} 
\caption{
    \textbf{Instance segmentation results comparing FPN~\cite{FPN} and FPN+FINE} on Mask R-CNN R50~\cite{MaskR-CNN} trained on COCO.
    Red solid boxes show zoomed-in views of the dashed regions.
}
\label{fig:appendix_segmentation}
\end{figure*}

\section{Additional Analysis}

\subsection{Qualitative Analysis}
\label{appendix_sec:Qualitative_Analysis}

\noindent\textbf{False Negative Reduction.}
As quantitatively shown in Table~\ref{tab:error_analysis}, the simultaneous reduction in both False Positives (FPs) and False Negatives (FNs) indicates that FINE not only suppresses background clutter but also amplifies foreground object signals.
As visualized in Figure~\ref{fig:appendix_FNs}, the baseline FPN frequently misses small or partially visible objects due to insufficient semantic abstraction in the low-level features. 
By enforcing semantic consistency across hierarchical levels, FINE recovers missed detections of small objects and those with low visual saliency (\eg, snowboards, frisbees, skis, and stop signs) without compromising correct predictions made by the baseline.

\noindent\textbf{Instance Segmentation.}
Figure~\ref{fig:appendix_segmentation} demonstrates that FINE's representational benefits extend to pixel-level instance segmentation. 
As shown in the red crops, the baseline FPN struggles with precise boundary localization, producing coarse masks that bleed into the background (e.g., teddy bear shadows) or fragment on complex shapes (\eg, skateboard edges). 
By modulating low-level features with high-level semantics, FINE resolves these pixel-level ambiguities, yielding cleaner instance separation and sharper boundaries.

\subsection{Detection Error Analysis on RT-DETR}
\label{subsec:error_analysis_RTDETR}

While FINE effectively suppresses False Positives (FPs) in region proposal-based detectors such as Mask R-CNN, as shown in Table~\ref{tab:error_analysis}, DETR-like architectures exhibit a distinct error profile.
As detailed in Table~\ref{tab:error_analysis_rtdetr_sweep}, integrating FINE into RT-DETR R18 consistently reduces False Negatives (FNs) and increases True Positives (TPs) across all fixed confidence thresholds.
However, this improvement in recall is accompanied by an apparent increase in FPs and a corresponding decrease in precision.

This behavior can be attributed to fundamental architectural differences.
Unlike Mask R-CNN, which explicitly filters background regions through the Region Proposal Network (RPN) and Non-Maximum Suppression (NMS), RT-DETR directly predicts a fixed set of object queries without NMS.
By enhancing foreground representations and suppressing background responses, FINE alters the confidence distribution of the predicted queries, generally shifting scores toward higher values.
This improves the confidence of true objects, reducing FNs, but can also increase the scores of ambiguous queries.
As a result, the F1-optimal operating threshold shifts upward (\eg, from $0.45$ to $0.55$).
At fixed, uncalibrated thresholds, this distributional shift causes more ambiguous queries to exceed the decision threshold, leading to an apparent increase in FPs (\eg, $+124\%$ at a $0.3$ threshold).

To disentangle FINE's actual detection effect from this calibration shift, we evaluate each model at its respective F1-optimal threshold (bottom row of Table~\ref{tab:error_analysis_rtdetr_sweep}).
Under this calibrated comparison, the apparent FP increase is reduced to $+4.8\%$ ($7,253$ to $7,603$).
Although precision decreases marginally from $0.743$ to $0.737$, recall improves from $0.571$ to $0.578$, resulting in an overall F1-score increase from $0.646$ to $0.648$.
These results suggest that, while FINE primarily improves \textit{precision} by suppressing background FPs in RPN-based models, its principal benefit in DETR-like architectures is improved object \textit{recall}.
Thus, the apparent FP increase observed at fixed thresholds is largely attributable to a shift in the confidence distribution rather than a degradation in predictive capability.

\begin{table}[!t]
\begin{center}
\resizebox{\columnwidth}{!}{
\begin{tabular}{clccccccc}
\toprule
Conf. & & \#Images & \#TPs ($\uparrow$) & \#FPs & \#FNs ($\downarrow$) & Precision & Recall & F1-score \\
\midrule
\multirow{2}{*}{0.3}
& w/o FINE         & 5,000 & 25,575 & \textbf{26,296} & 11,206 & 0.493 & 0.695 & 0.577 \\
& w/ FINE & 5,000 & \textbf{28,311} & 58,863 & \textbf{8,470}  & 0.325 & \textbf{0.770} & 0.457 \\
\midrule
\multirow{2}{*}{0.5}
& w/o FINE         & 5,000 & 19,597 & \textbf{4,930}  & 17,184 & 0.799 & 0.533 & 0.639 \\
& w/ FINE & 5,000 & \textbf{22,730} & 11,285 & \textbf{14,051} & 0.668 & \textbf{0.618} & \textbf{0.642} \\
\midrule
\multirow{2}{*}{0.7}
& w/o FINE         & 5,000 & 14,175 & \textbf{1,128} & 22,606 & \textbf{0.926} & 0.385 & 0.544 \\
& w/ FINE & 5,000 & \textbf{16,416} & 2,126 & \textbf{20,365} & 0.885 & \textbf{0.446} & \textbf{0.594} \\
\midrule
\multirow{2}{*}{F1-opt.}
& w/o FINE (0.45)  & 5,000 & 21,014 & \textbf{7,253} & 15,767 & \textbf{0.743} & 0.571 & 0.646 \\
& w/ FINE (0.55) & 5,000 & \textbf{21,260} & 7,603 & \textbf{15,521} & 0.737 & \textbf{0.578} & \textbf{0.648} \\
\bottomrule
\end{tabular}
}
\end{center}
\caption{
\textbf{Detection error analysis of RT-DETR R18 on COCO val2017 across confidence thresholds.}
Results are computed with an IoU threshold of $0.5$. 
``F1-opt.'' denotes the F1-optimal operating point for each respective model. 
}
\label{tab:error_analysis_rtdetr_sweep}
\end{table}




\end{document}